\documentclass[12pt]{article}
\usepackage[top=2.5cm, bottom=2.5cm, left=2.5cm, right=2.5cm]{geometry}

\usepackage{hyperref}
\usepackage{xcolor}
\usepackage{paralist,xspace,alltt,version}
\usepackage{stmaryrd}
\usepackage{listings}
\usepackage{pgfplots}
\usepackage{amsfonts,amssymb}
\usepackage{tikz}
\usetikzlibrary{calc,positioning,shapes,arrows,backgrounds}
\usepackage{dcolumn}
\usepackage{booktabs}
\usepackage{pdfpages}
\usepackage{amsmath}
\usepackage{mathabx}
\usepackage{rotating}
\usepackage[linewidth=1pt]{mdframed}
\usepackage{rotating}
\usepackage{pdflscape}
\usepackage{subfig}
\newcolumntype{M}[1]{D{.}{.}{1.#1}}

\newtheorem{theorem}{Theorem}
\newtheorem{definition}{Definition}
\newtheorem{assumption}{Assumption}
\newtheorem{example}{Example}

\newcommand{\beginproof}[1]{{\medskip\noindent {\bf Proof of {#1}:}\ }}
\newcommand{\QED}{\hfill\ensuremath{\blacksquare}} 
\newcommand{\true}{{\mathsf {true}}}
\newcommand{\Reals}{{\mathbb R}}
\newcommand{\Nat}{{\mathbb N}}
\newcommand{\means}[2]{{{#2}({#1})}}
\newcommand{\concrete}[1]{{\bigr [ \! \!\!\mid \! {#1}\! \mid \!\!\!\bigl ]}}
\newcommand{\tval}[1]{{{\{\mid {#1}\mid\}}}}
\newcommand{\maysat}{{\models^{{\small\mathsf{may}}}}}
\newcommand{\mustsat}{{\models^{{\small\mathsf{must}}}}}
\newcommand{\mayred}[2]{{\mathsf{Ex}({#1},{#2})}}
\newcommand{\union}[1]{{\,\stackrel{{\small {#1}}}{\cup}\,}}
\newcommand{\varmp}{{X_{mp}}}
\newcommand{\varx}{{X_x}}
\newcommand{\diff}{{\mathsf{diff}}}
\newcommand{\Sup}{{\mathsf{Sup}}}
\newcommand{\Inf}{{\mathsf{Inf}}}
\newcommand{\parents}[1]{{\mathsf {pnt}({#1})}}
\newcommand{\narrow}{}
\newcommand{\size}[1]{{\mid \! {#1} \! \mid}}
\newcommand{\wrap}{\star}

\begin{document}

\begin{center}
{\Large \bf Constrained Bayesian Networks:\\ Theory, Optimization, and Applications}\\

\bigskip
\bigskip
{\small Paul Beaumont and Michael Huth\\
Department of Computing, Imperial College London\\
London, SW7 2AZ, United Kingdom\\
$\{$paul.beaumont, m.huth$\}$@imperial.ac.uk}
\end{center}

\date{\today}

\bigskip
\begin{abstract}
We develop the theory and practice of an approach to modeling and
probabilistic inference in causal networks that is suitable when
application-specific or analysis-specific constraints should inform
such inference or when little or no data for the learning of causal
network structure or probability values at nodes are
available. Constrained Bayesian Networks generalize a Bayesian Network such that probabilities can be symbolic, arithmetic expressions and where the meaning of the network is constrained by finitely many formulas from the theory of the reals. 
A formal semantics for constrained Bayesian Networks over first-order
logic of the reals is given, which enables non-linear and non-convex optimization
algorithms that rely on decision procedures for this logic, and supports the composition of
several constrained Bayesian Networks. A
non-trivial case study in arms control, where few or no data are
available to assess the effectiveness of an arms inspection process,
evaluates our approach. An open-access prototype implementation of
these foundations and their algorithms uses the SMT solver Z3 as decision procedure, leverages an open-source package for Bayesian inference to symbolic computation, and is evaluated experimentally.
\end{abstract}


\section{Introduction}
\label{section:introduction}
Bayesian Networks (BN) \cite{p-bnmsmer-85,Pearl:1988:PRI:52121,Neapolitan:1990:PRE:77340} are a prominent, well established, and widely used formalism for expressing discrete probability distributions in terms of directed, acyclic graphs (DAG) that encode \emph{conditional} independence assumptions of distributions. Bayesian Networks have a wide range of applications~--~for example, trouble shooting \cite{DBLP:conf/uai/BreeseH96}, design of experiments \cite{DBLP:conf/psb/PageO06,DBLP:journals/expert/HarteminkGJY02}, 
and digital forensics to support legal reasoning
\cite{DBLP:conf/ifip11-9/TseCK12}. Their graph-based formalism and
automated support for probabilistic inference seem to lower adaption
hurdles for a diverse set of users with different technical
backgrounds. Bayesian Networks are also appealing since we may
combine, within the same Bayesian Network, different aspects such as
subjective beliefs expressed in probabilities, implicit trust
assumptions reflected in a bias of information processing or the
combinatorial logic of a process. Probabilistic inference for such
combinations is supported, including
belief updates based on observed evidence.  

Bayesian Networks also come with methodological support for learning
an appropriate graph structure as well as appropriate prior
probability values at nodes in such graphs from pre-existing data (see
for example
\cite{heckerman96,cussens15}). 
 The appropriateness of chosen prior probability values may depend on
 a variety of factors: the quality and quantity of data used for
 learning these values or the trust we place in experts who determine
 such values subjectively~--~to give two examples. We would therefore 
 like reassurance that the prior distributions represented by such
 values are robust enough in that small changes to such values only
 result in small changes of posterior distributions of interest. This
 naturally leads to the consideration of robust Bayesian statistics
 \cite{berger96,Berger2000}. 

A popular idea here is to approximate prior probabilities with
intervals and to then calculate~--~somehow~--~the intervals that
correspond to posterior probabilities. A good conceptual explanation
of this is Good's black box model \cite{good59,good82}, in which
interval information of priors is submitted into a black box that
contains all the usual methods associated with precise computations in
Bayesian Networks, and where the box then outputs intervals of
posteriors without limiting any interpretations or judgments on those output intervals. 

Our  engagement with a problem owner in arms control made us realize the benefits of Good's black box model and made us identify opportunities for extending it to increase the confidence that users from such problem domains can place in models and their robustness. Specifically, we want to be able to
\begin{itemize}
\item[R1] re-interpret compactly a BN as a possibly infinite set of
  BNs over the same graph, with robustness being analyzable over that re-interpretation

\item[R2] add logical constraints to capture domain knowledge or dependencies, and reflect constraints in 
robustness analyses in a coherent manner

\item[R3] compare models, within a composition context, to determine any differences in the robustness that they may offer for supporting decision making

\item[R4] parametrize the use of such a box so that it can produce outputs for any quantitative measure of interest definable as an arithmetic term

\item[R5] retain the ``blackness'' of the box so that the user neither has to see nor has to understand its inner workings

\item[R6] interpret outputs of the black box within the usual methodology of Bayesian Networks in as far as this may be possible.
\end{itemize}

\noindent We believe that these requirements are desired or apt in a
wide range of problem domains, in addition to the fact that they
should enhance usability of such a methodology in practice. We develop
\emph{constrained} Bayesian Networks in this paper and show that they
meet the above requirements. This development is informed by advances
made in areas of \emph{Symbolic Computation}
\cite{DBLP:journals/peerjpre/MeurerSPCRK0MSR16} and \emph{Automated
  Theorem Proving} \cite{DeMoura:2008:ZES:1792734.1792766} (with its applications to
\emph{Non-Linear, Non-Convex Optimization} \cite{DBLP:conf/esorics/BeaumontEHP15}), \emph{Three-Valued Model Checking} \cite{DBLP:conf/cav/BrunsG99}, and \emph{Abstract Interpretation}
\cite{DBLP:conf/csl/CousotC14}. 
Concretely,
we allow prior probabilities to be arithmetic expressions that may contain variables, and we enrich this model with logical constraints expressed in the theory of the reals. 

We draw comparisons to related work, including Credal Networks
\cite{DBLP:journals/ai/Cozman00,DBLP:conf/ijcai/CamposC05,Maua:2014:PIC:2693068.2693084}
and Constraint Networks \cite{dechter}. We then highlight similarities,
differences, and complementary value of our approach to this previous
work.

\paragraph{Contributions and methodology of the paper} We develop a formal syntax and
semantics of constrained Bayesian Networks which denote an empty,
finite or infinite set of Bayesian Networks over the same directed acyclic graph. We support this concept with a composition operator in which two or more constrained Bayesian Networks with different or ``overlapping'' graphs may be combined for cross-model analysis, subject to constraints that are an optional parameter of that composition. We formulate a three-valued semantics of the theory of the reals over constrained BNs that captures the familiar duality of satisfiability and validity but over the set of Bayesian Networks that a constrained Bayesian Network denotes. This semantics is used to reduce the computation of its judgments to satisfiability checks in the first-order logic over the reals. We then apply that reduction to design optimization algorithms that can compute, for any term definable in that logic,
infima and suprema up to a specified accuracy~--~for example for terms
that specify the meaning of marginal probabilities symbolically. These
optimization algorithms and their term parameter allow us to explore
or verify the robustness of a constrained Bayesian Network including,
but not limited to, the robustness of posterior distributions. We
demonstrate the use of such extended robustness analyses on a
non-trivial case study in the domain of arms control. We also
report a tool prototype that we have implemented and used to conduct these
analyses; it uses an SMT solver as a feasibility checker to implement these optimization algorithms; and it adapts an open-source package for Bayesian inference to symbolic computations.

Our principal theoretical contribution is the introduction of the
concept of a Constrained Bayesian Network itself, as well as its
intuitive yet formal semantics. Our theoretical results, such as those
for computational complexity and algorithm design, follow rather
straightforwardly from these definitions. This is because the latter
allow us to appeal directly to existing results from the existential
theory of the reals and optimization based thereupon.

Our main practical contribution is the successful integration of a
number of disparate techniques and approaches into a coherent semantic
framework and tool prototype that supports a range of modelling and
analysis capabilities, and does so in a highly automated manner.

\section{Background on Bayesian Networks}
\label{section:background}

A Bayesian Network (BN) is a graph-based statistical model for
expressing and reasoning about probabilistic uncertainty. The
underlying graph is directed and acyclic, a DAG. Nodes in this DAG
represent random variables defined by discrete probability
distributions that are also a function of the random variables
represented by the parent nodes in the DAG. In other words, a random
variable is conditioned on the random variables of its parent nodes.

We can use a BN to compute probabilities for events of interest over
these random variables. Bayesian inference also allows us to revise
such probabilities when additional observations of ``evidence'' have
been made.
\begin{figure}
\tikzset{
hollow node/.style={draw,ellipse,text width=2cm,align=center}
}

\hspace{-1cm}
\begin{center}
{\small
\begin{tikzpicture}[
node distance=0.8cm and 0cm,
mynode/.style={draw,ellipse,text width=2cm,align=center}
]

\node[mynode] (sp) {{\sf Holmes' Sprinkler}};
\node[mynode,below right=of sp] (gw) {{\sf Holmes' Grass Wet}};
\node[mynode,above right=of gw] (ra) {{\sf It Rains}};
\node[mynode,below right=of ra] (wa) {{\sf Watson's Grass Wet}};
\path
(ra) edge[-latex] (wa)
(sp) edge[-latex] (gw)
(ra) edge[-latex] (gw);

\node[left=of sp,left=0.3cm of sp] (sptab)
{
\begin{tabular}{M{1}M{1}}
\toprule
\multicolumn{2}{c}{{\sf Sprinkler}} \\
\multicolumn{1}{c}{On} & \multicolumn{1}{c}{Off} \\
\cmidrule{1-2}
0.1 & 0.9 \\
\bottomrule
\end{tabular}
};

\node[left=of sp,below=0.45cm of wa] (watab)
{
\begin{tabular}{cM{1}M{1}}
\toprule
& \multicolumn{2}{c}{{\sf Watson's}} \\
It Rains & \multicolumn{1}{c}{T} & \multicolumn{1}{c}{F} \\
\cmidrule(r){1-1}\cmidrule(l){2-3}
F & 0.05 & 0.95 \\
T & 0.7 & 0.3 \\
\bottomrule
\end{tabular}
};

\node[right=8.5cm of sptab] (ratab)
{
\begin{tabular}{M{1}M{1}}
\toprule
\multicolumn{2}{c}{{\sf It Rains}} \\
\multicolumn{1}{c}{T} & \multicolumn{1}{c}{F} \\
\cmidrule{1-2}
0.2 & 0.8 \\
\bottomrule
\end{tabular}
};

\node[below=0.2cm of sptab] (hotab)
{
\begin{tabular}{ccM{2}M{2}}
\toprule
& & \multicolumn{2}{c}{{\sf Holmes's}} \\
\multicolumn{2}{l}{Sprinkler/It Rains} & \multicolumn{1}{c}{T} & \multicolumn{1}{c}{F} \\
\cmidrule(r){1-2}\cmidrule(l){3-4}
Off & F & 0.05 & 0.95 \\
Off & T & 0.7 & 0.3 \\
On & F & 0.95 & 0.05 \\
On & T & 0.99 & 0.01 \\
\bottomrule
\end{tabular}
};

\end{tikzpicture}
} 
\end{center}

\caption{A BN with a 4-node DAG in the center and probability tables
  next to the respective nodes. This BN models beliefs about possible causes
  of wet grass on two neighbours' lawns. This BN allows us to reason
  about, for example, whether Holmes' lawn being wet is due to rain or
  Holmes' sprinkler~--~using observed evidence about a neighbour's lawn. \label{fig:simpleBN}}
\end{figure}
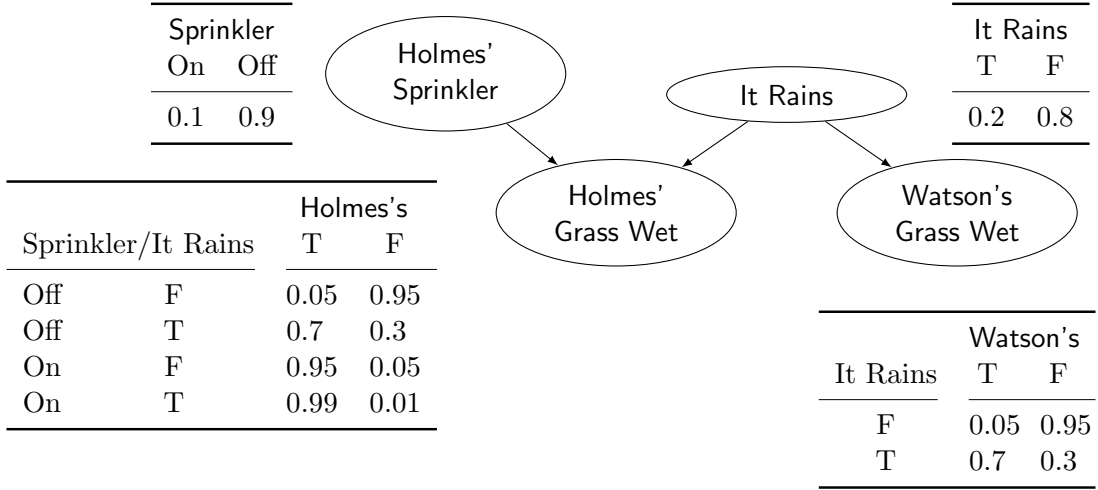

Figure~\ref{fig:simpleBN} shows a simple BN, which is part of the
folklore of example Bayesian Networks. It depicts the possible causes
of wet grass on two neighbours' lawns. 
For example, the probabilities of {\sf Holmes' Grass Wet} is
conditioned on its parents' output~--~whether {\sf It Rains} and {\sf
  Holmes' Sprinkler} is turned On or Off. The probability
of {\sf Holmes' Grass Wet} = T, given that {\sf Holmes' Sprinkler} =
  Off, Rain = F, for instance, is computed to be $0.05$, and is formally
stated as: 
\begin{equation}
p(\textnormal{{\sf Holmes' Grass Wet} = T} \mid \textnormal{{\sf Sprinkler} = Off, {\sf Rain} = F}) = 0.05 \nonumber
\end{equation}

This approach naturally gives cause to computations of the ``overall'' probability of an event happening, referred to as the \emph{marginal} probability. 
In the Bayesian approach, the Junction Tree Algorithm (JTA) (see e.g.\
Chapter~6 in \cite{barber12} for
further details) 
may be used to revise a marginal of a BN because of
``hard'', respectively ``soft'', evidence~--~the \emph{definite},
respectively \emph{probabilistic}, observation of an additional or
new event. 

We now formalize BNs and use this below to enrich
BNs with modeling and reasoning capabilities that realize the aforementioned requirements.
\begin{definition}
\begin{enumerate}
\item A \emph{Bayesian network (BN)} is a pair $(G,\pi)$ where $G$ is
  a finite, directed, acyclic graph $G= (N,E)$ of nodes $N$ and edge
relation $E\subset N\times N$, and where $\pi$ is a tuple
$(\pi_n)_{n\in N}$ of formal probability tables. 

\item The formal probability table $\pi_n$ is defined as follows. 
Let $\parents n = \{ n'\in N\mid (n',n)\in E\}$ be
  the (possibly empty) set of parents of $n$ in DAG $G$ and $O_n$ the
  set of outcomes of the random variable at node $n$.
Then $\pi_n$ is a discrete probability distribution, a function $\pi_n$ of
type \(\bigl (\prod_{n'\in\parents n} O_{n'}\bigr )\times O_n\to
[0,1]\) such that its mass $\sum_e \pi_n(e)$ equals $1$.

\end{enumerate}
\end{definition}

Above, it is understood that $\prod_{n'\in\emptyset} O_{n'}$ equals
$\{*\}$; in that case, $\pi_n$ has type isomorphic to $O_n\to [0,1]$.

\section{Constrained Bayesian Networks}
\label{section:theory}
Informally, a constrained BN
is obtained from a BN by replacing one or
more probabilities in its probability tables with symbolic
expressions, and by adding constraints for variables used
in these expressions or in quantitative terms of interest,
and for variables that refer to marginal
probabilities of interest. We write $B^C_X$ for constrained BN
with set of constraints $C$ and variable set $X$.

To illustrate, in Figure~\ref{fig:cBNexample} the
probability tables for two nodes {\sf Sprinkler} and {\sf  Rain} of the BN in Figure~\ref{fig:simpleBN}
are made symbolic with a variable $x$ to obtain a constrained BN.
This allows us to model \emph{strict uncertainty} (also known as
Knightian uncertainty) in the actual value of such probabilities. Our approach allows variables to
be shared across such tables, as $x$ is shared across the tables for
{\sf Sprinkler} and {\sf Rain}. This is certainly useful, e.g., 
to express that a certain subjective probability is twice as likely as
another one. 

We use variables $mp_H$ and
$mp_W$ to refer to marginal probabilities
\begin{eqnarray}
p(\textnormal{{\sf Holmes' Grass Wet} = True}) &{}& {}\label{equ:Holmes}\\
p(\textnormal{{\sf Holmes' Grass Wet} = True}\mid \textnormal{{\sf Watson's\
  Grass\ Wet} = True}) {} &{}& {}\label{equ:Watson}
\end{eqnarray}

\noindent respectively. The constraints we then consider are
$0\leq x\leq 1$, to ensure that symbolic expressions still specify probability distributions, as well as the symbolic meaning of the marginal
probabilities $mp_H$ and $mp_W$ which are captured in two \emph{non-linear} 
equations in
$x$ as
\begin{eqnarray}
{} &{}& mp_H = 
0.495\! * \! x\! * \! x + 0.5\! * \! x\! * \! (-0.95\! * \! x + 0.95) + \nonumber\\
{} &{}& 0.7\! * \! x\! * \! (-0.5\! * \! x + 1) + 1.0\! * \! (-0.5\! * \! x +
1)\! * \! (-0.05\! * \! x + 0.05) \label{equ:mpH}
\end{eqnarray}
\begin{eqnarray}
{} &{}& mp_W \! * \!   (0.35\! * \! x\! * \! x + 0.025\! * \! x\! * \! (-0.95\! * \! x + 0.95) + 0.7\! * \! x\! * \! (-x\! * \! 0.5 + 1) +\nonumber\\
{} &{}& 0.025\! * \! x\! * \! (-0.05\! * \! x + 0.05) + 0.05\! * \! (-0.95\! * \! x + 0.95)\! * \! (-x\! * \! 0.5 + 1) +\nonumber\\
{} &{}& 0.05\! * \! (-x\! * \! 0.5 + 1)\! * \! (-0.05\! * \! x + 0.05)) = \nonumber\\
{} &{}& (0.3465\! * \! x\! * \! x + 0.025\! * \! x\! * \! (-0.95\! * \! x + 0.95) + 0.49\! * \! x\! * \! (-x\! * \! 0.5 + 1) + \nonumber\\
{} &{}& 0.05\! * \! (-x\! * \! 0.5 + 1)\! * \! (-0.05\! * \! x + 0.05))\label{equ:mpW}
\end{eqnarray}
\begin{figure}
\begin{center}

\tikzset{
hollow node/.style={draw,ellipse,text width=2cm,align=center}
}

\begin{center}

\begin{tikzpicture}[
node distance=1cm and 2cm,
mynode/.style={draw,ellipse,text width=2cm,align=center}
]

\node[] (sptab)
{
\begin{tabular}{M{1}M{1}M{1}M{1}M{1}}
\toprule
\multicolumn{5}{c}{{\sf Sprinkler}} \\
\multicolumn{2}{c}{On} & & \multicolumn{2}{c}{Off} \\
\cmidrule{1-5}
& 0.5*x & & & 1-0.5*x \\
\bottomrule
\end{tabular}
};

\node[right=of sptab] (ratab)
{
\begin{tabular}{M{1}M{1}M{1}}
\toprule
\multicolumn{3}{c}{{\sf Rain}} \\
\multicolumn{1}{c}{T} &  \multicolumn{2}{c}{F} \\
\cmidrule{1-3}
x & & 1-x\\
\bottomrule
\end{tabular}
};

\end{tikzpicture}

\end{center}

\end{center}
\caption{Probability tables that make a constrained BN $B^{C_0}_{X_0}$
  out of the BN $B$ in Figure~\ref{fig:simpleBN}
  by entering non-constant terms into tables {\sf Sprinkler}
  and {\sf Rain}. Marginal probabilities of interest are those given
  in~(\ref{equ:Holmes}) and~(\ref{equ:Watson}) and their meaning is defined as constraints in~(\ref{equ:mpH}) and~(\ref{equ:mpW})\label{fig:cBNexample}}
\end{figure}

The above equations are constructed through symbolic interpretations of computations of marginals, for example of the Junction Tree Algorithm, and subsequent elimination of division operators. The latter computes a normal form of rational terms from which these equations are easily derived.

\subsection{Theoretical Foundations}
We begin the formal development by defining grammars for symbolic expressions that occur in
probability tables and for properties that contain such expressions as arguments.
Figure~\ref{fig:grammars} shows definitions for real-valued terms $t$,
where $c$ ranges over real constants, and $x$ and $mp$ are real variables ranging over
variable sets $\varx$ and $\varmp$, respectively. The distinction
between $mp$ and $x$ is one of modelling intent. Variables $mp$ refer
to marginal probabilities of a constrained BN $B^C_X$. The meaning of
these \emph{symbolic} marginals is defined via constraints in $C$.
Variables in $\varx$ may occur in symbolic expressions in
probability tables of nodes or denote any quantitative measures of
interest.
We write
\(X = \varx\cup \varmp\)
for the disjoint union of such variable sets.

Constraints $\varphi$ are quantifier-free formulas
built from inequalities over terms $t$, logical truth constant $\true$, and
propositional operators. Queries $\phi$ are built out of constraints
and first-order quantifiers.
\begin{definition}
We write $\mathcal T[X]$ for the set of all terms $t$, $\mathcal C[X]$ for
the set of all constraints $\varphi$ generated in this manner, and we
write $\mathcal Q[X]$ for the set of all queries $\phi$ generated in
this manner from variable set $X$.
\end{definition}

\noindent We write $\mathcal T$, $\mathcal C$, and $\mathcal Q$
whenever $X$ is clear from context and write $\lor$,
$-$, $=$, $>$ and so forth for derived logical, arithmetic, and relational operators. 
\begin{figure}
\(t ::= c\ \mid\ mp\ \mid\ x\ \mid\ t+t\ \mid t*t\)

\(\varphi ::= \true\ \mid\ t\leq t\ \mid\ t < t\ \mid \lnot\varphi\ \mid\ \varphi\land\varphi\)

\(\phi ::= \varphi\ \mid \exists x\colon \phi\ \mid\ \lnot\phi\ \mid \phi\land\phi\)
\caption{BNF grammars for real-valued terms $t$, constraints $\varphi$, and queries $\phi$ 
where $x$ are variables from set $\varx$, $c$ are constant reals, $mp$ are variables from set $\varmp$ denoting marginal probabilities, and $\true$ denotes logical truth\label{fig:grammars}}
\end{figure}

We may think of a \emph{constrained} BN $B^C_X$ as a BN $B$ in which
entries in probability tables of nodes may not only be concrete
probabilities but terms $t$ of the grammar in
Figure~\ref{fig:grammars} over variable set $\varx$, and where the BN is
enriched with a finite set of constraints $C = \{\varphi_i\mid 1\leq
i\leq n\}$. The intuition is that $B^C_X$ denotes a \emph{set of BNs
  that all have the same graph and the same structure of probability
  tables} but where \emph{probability values may be uncertain, 
  modelled as arithmetic terms, and subject to application-specific
  or analysis-specific constraints}. 
The only difference in two BNs from
that set may be in the real number entries in those probability
tables, and those real numbers are instantiations of the specified
arithmetic terms such that all constraints are met. We formalize this:
\begin{definition}
A constrained BN of type $(\varx, \varmp)$~--~denoted as
$X=\varx\cup\varmp$ by abuse of notation~--~is a triple $(G,C,\pi)$
where $G=(N,E)$ is a finite DAG, $C$ a finite set of constraints
from $\mathcal C[X]$, and $\pi$ a tuple $(\pi_n)_{n\in N}$ of
\emph{symbolic} probability tables with $O_n$ as the set of outcomes
of random variable at node $n$:
\begin{equation}
\pi_n\colon \bigl (\prod_{n'\in\parents n} O_{n'}\bigr )\times O_n\to
\mathcal T[\varx] \nonumber
\end{equation}
\end{definition}

Note that a symbolic probability table has the same input type as a
formal probability table, but its output type is a set of terms not
the unit interval. 
Let us first define syntactic restrictions for constrained BNs.
\begin{definition}
\begin{enumerate}
\item A constrained BN $(G,C,\pi)$ of type $X$ is
\emph{well-formed} if
\begin{enumerate}
\item $X = \varx\cup \varmp$ equals the set of variables that occur in $C$

\item all $mp$ in $\varmp$ have exactly one defining
  equation $mp = t$ or $mp*t = t'$ in $C$ where neither $t$ nor $t'$
  contain variables from $\varmp$.
\end{enumerate}

\item When $G$ and $\pi$ are determined by context, we 
refer to a well-formed, constrained BN
  $(G,C,\pi)$ of type $X$ as $B^C_X$.
\end{enumerate}
\end{definition}

Item~1(a) says that all variables in $X$ occur in some constraint from $C$. Item~1(b) ensures all variables $mp$ that model marginal probabilities have a defined meaning in $C$.
Note that item~1(b) is consistent with having other constraints on such variables in $C$, for example a constraint saying that $0.1\leq mp\leq x*y$. 
These items
create a two-level term language,
with variables in $\varx$ informing meaning of variables in $\varmp$.

A sound, constrained BN has a \emph{semantic} requirement
about its \emph{concretizations}, which we now formalize using
assignments for quantifier-free formulas.
\begin{definition}
\begin{enumerate}
\item An assignment $\alpha$ is a function $\alpha\colon
  X\to \Reals$. For $c$ in $\Reals$ and $x$ in $X$, assignment $\alpha[x\mapsto c]$ equals $\alpha$ except at  $x$, where it outputs $c$.

\item The meaning $\means t\alpha$ of term $t$ in $\mathcal T$ under $\alpha$, as well as the  judgment $\alpha\models \phi$ for all $\phi$ in
  $\mathcal Q$, are defined in Figure~\ref{fig:termsformulas}.
\end{enumerate}
\end{definition}
\begin{figure}
{\small
\baselineskip=8pt
\begin{eqnarray*}
\means {c}\alpha &=& c \qquad\qquad \hbox{constants denote themselves}\\
\means {x}\alpha &=& \alpha(x) \qquad\qquad \hbox{for all $x$ in $X$,
  overloading of notation}\\
\means {t_1+t_2}\alpha &=& \means {t_1}\alpha + \means {t_2}\alpha\\
\means {t_1*t_2}\alpha &=& \means {t_1}\alpha * \means {t_2}\alpha\\
{} &{}& {}\\
\alpha \models\true &{}& \hbox{holds}\\
\alpha\models t_1\leq t_2 &\hbox{iff}& \means {t_1}\alpha \leq \means {t_2}\alpha\\
\alpha \models t_1 < t_2 &\hbox{iff}& \means {t_1}\alpha < \means {t_2}\alpha\\
\alpha \models \lnot\varphi &\hbox{iff}& (\hbox{not } \alpha \models \varphi)\\
\alpha \models \varphi_1\land\varphi_2 &\hbox{iff}& (\alpha \models
\varphi_1 \hbox{ and } \alpha \models \varphi_2)\\
\alpha \models \exists x\colon \phi &\hbox{iff}& \alpha[x\mapsto c]\models \phi\hbox{ for some real number } c\\
\alpha \models \lnot\phi &\hbox{iff}& (\hbox{not } \alpha\models \phi)\\
\alpha\models \phi_1\land\phi_2 &\hbox{iff}& (\alpha\models\phi_1 \hbox{ and } \alpha\models \phi_2)
\end{eqnarray*}
} 
\caption{Top: meaning of terms for an assignment $\alpha$, where we identify constants with their meaning, and use the same symbol $+$ for syntax and semantics, and similarly for $*$. Bottom: Semantics of judgment $\alpha\models\phi$ for an assignment
$\alpha\colon X\to \Reals$ and $\phi$ from $\mathcal Q$. This uses the meaning $\means {t}\alpha$ from Figure~\ref{fig:termsformulas} and uses $\leq$ both as syntax and semantics, similarly for $<$\label{fig:termsformulas}}
\end{figure}

\noindent Note that $\means t\alpha$ extends
$\alpha\colon X\to \Reals$ to type $\mathcal T[X]\to \Reals$.
The judgment $\alpha\models \phi$ is satisfaction of
first-order logic over the reals. 
We use these judgments to define the set of
concretizations of a well-formed, constrained BN:
\begin{definition}
Let $B^C_X = (G,C,\pi)$ be a well-formed, constrained BN
where $G=(N,E)$. Let $\alpha\colon X\to \Reals$ be an assignment.
\begin{enumerate}
\item  We write $B^C_X[\alpha]$ for
  the BN $(G,\pi[\alpha])$ that forgets $C$ from $B^C_X$ and
  has formal probability table $\pi[\alpha]_n$ for each node $n$ with
\(\pi[\alpha]_n = \lambda e\colon \means {\pi_n(e)}\alpha \nonumber\).
 
\item The set $\concrete {B^C_X}$ of BNs that $B^C_X$ denotes, its set of \emph{concretizations}, is
\begin{equation}
\label{equ:meaning}
\concrete {B^C_X} = \{ B^C_X[\alpha]\mid \alpha\colon X\to \Reals \hbox{
  and } \alpha\models \bigwedge_{\varphi'\in C}\varphi'\}
\end{equation}
\end{enumerate}
\end{definition}

Note that the formal probability
table $\pi[\alpha]_n$ computes $\pi[\alpha]_n(e)$ as
$\means t\alpha$ where $t$ is the term $\pi_n(e)$ in $\mathcal T[\varx]$.
We can now define \emph{sound} constrained BNs.
\begin{definition}
Let $B^C_X = (G,C,\pi)$ be a well-formed, constrained BN. Then
$B^C_X$ is \emph{sound} if for all $B^C_X[\alpha]$ that are
concretizations of $B^C_X$ we have, for all nodes $n$ and inputs $e$ of
$\pi[\alpha]_n$, that $\pi[\alpha]_n(e)$ is in $[0,1]$ and \(\sum_e \pi[\alpha]_n(e) = 1\).
\end{definition}

Soundness is saying that all concretizations of a well-formed, constrained BN are actually BNs: for each such $B^C_X[\alpha]$ and
node $n$ in it, $\pi[\alpha]_n$ is a discrete probability distribution.
\begin{assumption}
All constrained BNs used in this paper are sound.
\end{assumption}

It is important to know whether $\concrete {B^C_X}$ is non-empty. 
\begin{definition}
A constrained $B^C_X$ is consistent iff $\concrete {B^C_X}\not=\emptyset$.
\end{definition}

The techniques developed in the next Section~\ref{section:semjud} will also allow us to decide whether a constrained BN is consistent.

\subsection{Semantic Judgments}
\label{section:semjud}
How should we best reason about a set of BNs $\concrete {B^C_X}$? We
propose two semantic judgments that allow us to explore worst-case and
best-case properties of $B^C_X$. A judicious combination of these
judgments also enables us to express optimizations over the
imprecision and probabilistic uncertainty inherent in $B^C_X$, whilst
reflecting any application-specific or analysis-specific constraints. Both semantic judgments rest on a satisfaction relation between concretization BNs and queries. We define this formally.
\begin{definition}
Let $B^C_X$ be a constrained BN. For all $\phi$ in $\mathcal Q$, the two semantic judgments $\mustsat$ and $\maysat$ are defined as
\begin{eqnarray}
B^C_X \mustsat \phi &\hbox{iff}&\hbox{for all $B^C_X[\alpha]$ in $B^C_X$
  we have $\alpha \models \phi$}\label{equ:mustsat}\\
B^C_X \maysat \phi &\hbox{iff}&\hbox{for some $B^C_X[\alpha]$ in $B^C_X$
  we have $\alpha \models\phi$}\label{equ:maysat}
\end{eqnarray}
\end{definition}

\noindent The definition in~(\ref{equ:mustsat}) allows us to discover \emph{invariants}: truth of $B^C_X\mustsat\phi$ implies that $\phi$ holds no matter what concrete instance in $\concrete {B^C_X}$ the modeller may face, a form of worst-case reasoning. Dually, the truth of $B^C_X\maysat\phi$ in~(\ref{equ:maysat}) implies it is possible that the modeller faces a BN in $\concrete {B^C_X}$ that satisfies $\phi$, a form of best-case reasoning. We may formalize this duality.
\begin{theorem}
\label{theorem:dual}
For all constrained BNs $B^C_X$ and $\phi$ in $\mathcal Q$ we have
\begin{enumerate}
\item $B^C_X\mustsat\phi$ iff not $B^C_X\maysat\lnot\phi$
\item $B^C_X\maysat\phi$ iff not $B^C_X\mustsat\lnot\phi$
\item  $B^C_X\mustsat\phi_1\land\phi_2$ iff ($B^C_X\mustsat\phi_1$ and $B^C_X\mustsat\phi_2$)
\item  $B^C_X\maysat\phi_1\lor\phi_2$ iff ($B^C_X\maysat\phi_1$ or $B^C_X\maysat\phi_2$)
\end{enumerate}
\end{theorem}

We illustrate this formalization of constrained BNs with an example.

\begin{example}
\label{example:cBN}
The constrained BN from Figures~\ref{fig:simpleBN}
and~\ref{fig:cBNexample} is sound and 
consistent. Consider the query $\varphi_H$ being $mp_H < 0.3$ where we use variable $mp_H$ to denote the marginal probability $p(\textnormal{{\sf Holmes' Grass Wet} =
True})$. We mean to compute whether
$B^C_X\mustsat \varphi_H$ and $B^C_X\maysat \varphi_H$ hold for this
constrained BN. We conclude that $B^C_X\mustsat \varphi_H$ does
not hold: $p(\textnormal{{\sf Holmes' Grass Wet} = True})$ equals $0.35255$ 
for $\alpha(x)=0.3$ and so $\alpha(mp_H) = 0.35255$ as well, and we have
$\alpha\models \bigwedge_{\varphi'\in C} \varphi'$ and
$0.35255\not\leq 0.3$. On the other hand, $B^C_X\maysat \varphi_H$
holds since for $\alpha'(x)=0.1$ we have $\alpha'\models \bigwedge_{\varphi'\in C} \varphi'$ and $\alpha'(mp_H) = 0.15695$ is less than or equal to $0.3$.  

Observing additional hard evidence that Watson's grass is
wet, we similarly evaluate judgments $B^C_X\maysat\varphi$
and $B^C_X\mustsat\varphi$ when $\varphi$ contains $mp_W$ which refers
to marginal
\[p(\textnormal{{\sf Holmes' Grass Wet} = True}\mid \textnormal{{\sf Watson's\
  Grass\ Wet} = True})
\]

\end{example}

\subsection{Consistent constrained BNs}
It is important to understand how the semantic judgments $\maysat$ and
$\mustsat$ relate to consistent or inconsistent constrained BNs. We
can characterize consistency through properties of these
semantic judgments:
\begin{theorem}
\label{theorem:conchar}
Let $B^C_X$ be a constrained BN. Then the following are all equivalent:
\begin{enumerate}
\item $B^C_X\maysat\true$ holds.

\item $B^C_X$ is consistent.

\item For all $\phi$ in $\mathcal Q$, we have that $B^C_X\mustsat\phi$ implies $B^C_X\maysat \phi$.

\item For all $\phi$ in $\mathcal Q$, we have that $B^C_X\maysat
  \phi\lor\lnot\phi$ holds.

\item For all $\phi$ in $\mathcal Q$, we have that $B^C_X\mustsat
  \phi\land\lnot\phi$ does not hold.
\end{enumerate}
\end{theorem}

We stress that it is vital to check the consistency of $B^C_X$ prior
to relying on any findings of its further analysis. If $B^C_X$ is
inconsistent, then $\concrete {B^C_X}$ is empty and so
$B^C_X\mustsat\phi$ holds trivially for all $\phi$ in $\mathcal Q$
since the universal quantification of its defining semantics
in~(\ref{equ:mustsat}) ranges over the empty set. Not detecting such
inconsistency may thus lead to unintended and flawed reasoning. In
our tool, this is a non-issue as it uses these judgments within
optimization algorithms that either report a concretization as witness
or report a discovered inconsistency.

Consistency checking is NP-hard: checking the satisfiability of constraints in logic $\mathcal C$ is NP-hard. And so this hardness is inherited for any notion of size of a constrained BN that includes the sum of the sizes of all its constraints.

\subsection{Reducing $\maysat$ and $\mustsat$ to satisfiability checking}
Our case studies involving constrained BNs suggest that it suffices to
consider elements of $\mathcal C$, i.e.\ to consider
formulas of $\mathcal Q$ that are quantifier-free. The benefit of
having the more expressive logic $\mathcal Q$, however, is that its
quantifiers allow us to reduce the decisions for $B^C_X\maysat \varphi$
and $B^C_X\mustsat \varphi$ for quantifier-free formulas $\varphi$ to
satisfiability checking, respectively validity checking in the logic
$\mathcal Q$~--~which we now demonstrate. 
For sound, constrained BNs $B^C_X$, the judgment $B^C_X\maysat \varphi$ asks whether there is an
assignment \(\alpha\colon X \to \Reals\)  such that
$\alpha\models \bigwedge_{\varphi'\in C} \varphi'$ and $\alpha\models\varphi$ both hold. Since $\mathcal C$ is
contained in $\mathcal Q$, we may capture this meaning within the logic
$\mathcal Q$ itself as a satisfiability check. 
Let set $X$ equal $\{x_1,\dots, x_n\}$.
Then, asking whether $\alpha\models \bigwedge_{\varphi'\in C} \varphi'$ and $\alpha\models\varphi$ hold,
asks whether the formula in~(\ref{equ:maysatreduction}) of logic $\mathcal Q$ is satisfiable:
\begin{equation}
\label{equ:maysatreduction}
\exists x_1\colon \dots \colon \exists x_n\colon \varphi\land \bigwedge_{\varphi'\in C}\varphi'
\end{equation}
\begin{definition}
For a constrained BN $B^C_X$ and $\varphi$ in $\mathcal C$, we write
$\mayred{B^C_X}{\varphi}$ to denote the formula defined
in~(\ref{equ:maysatreduction}). 
\end{definition}

\noindent Note that $\mayred{B^C_X}{\varphi}$
depends on $B^C_X$: namely on its set of variables $X$ and constraint
set $C$, the latter reflecting symbolic meanings of marginal probabilities. Let us illustrate this by revisiting Example~\ref{example:cBN}.
\begin{example}
\label{holmes}
For $\varphi_H$ as in Example~\ref{example:cBN} with type $X =
\{x\}\cup \{mp_H\}$, the formula we derive for $B^C_X\maysat \varphi_H$ is
\begin{eqnarray}
{} &{}& \exists mp_H\colon \exists x\colon (mp_H < 0.3)\land (0.1\leq x)\land (x\leq 0.3)\land \label{equ:expH}\\
{} &{}& \qquad\qquad\ \ \  (0.5*x + (1-0.5*x) = 1)\land (x+(1-x) = 1)\land (mp_H = t)\nonumber
\end{eqnarray}

\noindent where $t$ is the term on the righthand side of the equation in~(\ref{equ:mpH}).
\end{example}

We can summarize this
discussion, where we also appeal to the first
item of Theorem~\ref{theorem:dual} to get a similar characterization
for $\mustsat$.
\begin{theorem}
\label{theorem:reduction}
Let $B^C_X$ be a constrained BN and $\varphi$ in $\mathcal C$.
Then we have:
\begin{enumerate}
\item Formula $\mayred{B^C_X}\varphi$ in~(\ref{equ:maysatreduction}) is in $\mathcal Q$ and in the existential fragment of $\mathcal Q$. 

\item Truth of
$B^C_X\maysat\varphi$ is equivalent to the satisfiability of $\mayred{B^C_X}\varphi$ in $\mathcal Q$.

\item $B^C_X\maysat\varphi$ can be decided in PSPACE in the size of formula $\mayred{B^C_X}{\varphi}$.

\item $B^C_X\mustsat\varphi$ can be decided in PSPACE in the size of formula $\mayred{B^C_X}{\lnot \varphi}$.
\end{enumerate}
\end{theorem}

This result of deciding semantic judgments in polynomial space 
pertains to the size of formulas
$\mayred{B^C_X}{\varphi}$ and $\mayred{B^C_X}{\lnot \varphi}$, and
these formulas contain equations that define the meaning of marginals
symbolically. There is therefore an incentive to simplify such
symbolic expressions prior to their incorporation into $C$ and these formulas, and we do such simplifications in our implementation.

\subsection{Constrained Union Operator}
\label{subsection:union}
We also want the ability to compare two or more constrained BNs or to discover relationships between them. This is facilitated by a notion of composition of constrained BNs, which we now develop. Consider two constrained BNs $B^{C_1}_{X_1}$ and $B^{C_2}_{X_2}$. 
Our intuition for composition is to use a \emph{disjoint union} of the graphs of
each of these constrained BNs such that each node in this unioned DAG
still has its symbolic probability table as before. 
This union operator renames nodes that appear in both graphs so that
the union is indeed disjoint.
As a set of constraints for the resulting constrained BN, we then
consider $C_1\cup C_2$. 

It is useful to make this composition depend on another set of constraints $C$. The idea is that $C$ can specify known or assumed relationships between these BNs. The resulting composition operator $\union C$ defines the composition
\begin{equation}
\label{equ:union}
B^{C_1}_{X_1}\union C B^{C_2}_{X_2}
\end{equation}

\noindent as the constrained BN with graph and probability tables
obtained by disjoint union of the graphs and symbolic probability tables of
$B^{C_1}_{X_1}$ and $B^{C_2}_{X_2}$, where the set of constraints for
this resulting constrained BN is now $C_1\cup C_2\cup C$.

 This composition operator has an implicit assumption for being well
 defined, namely that $C$ does not contain any equations that
 (re)define the (symbolic) meaning of marginal probabilities given in
 $C_1\cup C_2$. 
\begin{figure}
\tikzset{
hollow node/.style={draw,ellipse,text width=2cm,align=center}
}
\hspace{-0.1cm}
\begin{center}
{\small
\begin{tikzpicture}[
node distance=0.6cm and 0cm,
mynode/.style={draw,ellipse,text width=1.5cm,align=center}
]

\node (p4) {} ;

\node[above=0.1cm of p4] (sptab2)
{
{\small
\begin{tabular}{M{1}M{1}M{1}M{1}M{1}}
\toprule
\multicolumn{4}{c}{{\sf Sprinkler}} \\
\multicolumn{2}{c}{On}  & \multicolumn{2}{c}{Off} \\
\cmidrule{1-4}
& z & & 1-z \\
\bottomrule
\end{tabular}
} 
};

\node[below=0.1cm of p4] (ratab)
{
{\small
\begin{tabular}{M{1}M{1}M{1}M{1}}
\toprule
\multicolumn{4}{c}{{\sf Rain}} \\
\multicolumn{1}{c}{Heavy} & \multicolumn{1}{c}{Light} &  \multicolumn{2}{c}{None}  \\
\cmidrule{1-4}
y & 4y & & 1-5y \\
\bottomrule
\end{tabular}
} 
};

\node[ right=2.1cm of p4] (hotab)
{
{\small
\begin{tabular}{ccM{2}M{2}}
\toprule
& & \multicolumn{2}{c}{{\sf Holmes'}} \\
\multicolumn{2}{l}{Sprinkler/Rain} & \multicolumn{1}{c}{T} & \multicolumn{1}{c}{F} \\
\cmidrule(r){1-2}\cmidrule(l){3-4}
Off & N & 0.05 & 0.95 \\
Off & L & 0.65 & 0.35 \\
Off & H & 0.9 & 0.1 \\
On & N & 0.95 & 0.05 \\
On & L & 0.95 & 0.05 \\
On & H & 0.99 & 0.01 \\
\bottomrule
\end{tabular}
} 
};

\node[right=0.1cm of hotab] (watab)
{
{\small
\begin{tabular}{cM{1}M{1}}
\toprule
& \multicolumn{2}{c}{{\sf Watson's}} \\
Rain & \multicolumn{1}{c}{T} & \multicolumn{1}{c}{F} \\
\cmidrule(r){1-1}\cmidrule(l){2-3}
N & 0.05 & 0.95 \\
L & 0.65 & 0.35 \\
H & 0.9 & 0.1 \\
\bottomrule
\end{tabular}
} 
};

\end{tikzpicture}
} 
\end{center}
\caption{
  Symbolic probability tables for a constrained BN
  $B^{C'_0}_{X'_0}$ that has the same DAG as $B^{C_0}_{X_0}$ of
  Figure~\ref{fig:cBNexample} but is more complex: rain is modelled
  to a greater degree of specificity. 
Variable set $X'_0$ is $\{y,z\}\cup \{mp'_H,
  mp'_W\}$. The constraint set $C'_0$ includes $0.1\leq 5*y\leq 0.3$,
  equations that define
  the meaning of marginals $mp'_W$ and $mp'_W$ in terms of $y$ and
  $z$ (not shown), and equations that ensure that all tables specify
  probability tables.   \label{fig:morecomplicatedBBN}}
\end{figure}

We give an example of such a union of constrained BNs that already
illustrates some reasoning capabilities to be developed in this paper:
\begin{example}
\label{examplediff}
Figure \ref{fig:morecomplicatedBBN} specifies a constrained BN
$B^{C'_0}_{X'_0}$ that is similar to constrained BN $B^{C_0}_{X_0}$
defined in Figure~\ref{fig:cBNexample} but that models rain with more
specificity. Variables $y$ and $z$ are used in symbolic probabilities,
and variables $mp'_H$ and $mp'_W$ refer to the marginals
in~(\ref{equ:Holmes}) and~(\ref{equ:Watson}) respectively. The
constraint $0.1\leq 5*y\leq 0.3$ in $C'_0$ corresponds to the
constraint $0.1\leq x\leq 0.3$ in $C_0$ and so term $5*y$ in some way
reflects $x$, that it rains according to $B^{C_0}_{X_0}$. 

The constraint set $C$ that binds the two models together is $\{2*z =
x\}$, which ensures that the probability for the sprinkler to be on is
the same in both models. In the constrained BN $B^{C_0}_{X_0}\union C
B^{C'_0}_{X'_0}$, we want to understand the difference in the marginal
probabilities $mp_W$ and $mp'_W$, expressed by term $\diff =
mp_W-mp'_W$. 

Subtraction $-$ and equality $=$ are derived operations in $\mathcal Q$.  The methods we will develop in this paper allow us to conclude that
the maximal value of $\diff$ is in the closed interval 
$[0.134079500198,0.134079508781]$, with $\diff$ being
$0.134079500198$ when
\[
\begin{array}{lll}
 x = 0.299999999930 &  z = 0.149999999965 & y = 0.020000000003 \\
mp_W = 0.663714285678 &  mp'_W = 0.529634782614 & {}
\end{array}
\]

\noindent Similarly, we may infer that the minimum of
$\diff$ is in the closed interval
\[ 
[-0.164272228181,-0.164272221575]
\]

\noindent with $\diff$ being $-0.164272221575$ when
\begin{equation}
\begin{array}{lll}
x = 0.100000000093 & z = 0.050000000046 & y = 0.059999999855 \\ 
mp_W = 0.472086956699 & mp'_W = 0.636359183424 & {} 
\end{array}
\nonumber
\end{equation}

\noindent In particular, the absolute value of the difference of the
marginal probability~(\ref{equ:Watson}) in those constrained BNs
is less than $0.1643$, attained for the values just shown.
\end{example}

These union operators are symmetric
in that $B^{C_1}_{X_1}\union C B^{C_2}_{X_2}$ and $B^{C_2}_{X_2}\union
C B^{C_1}_{X_1}$ satisfy the same judgments $\mustsat$ and
$\maysat$ for all $\phi$ in $\mathcal Q$. Idempotency won't hold in
general as unions may introduce a new set of constraints $C$. Associativity holds,
assuming all compositions in~(\ref{equ:assoc}) give rise to sound constrained BNs:
\begin{equation}
\label{equ:assoc}
(B^{C_1}_{X_1}\union C B^{C_2}_{X_2})\union {C'} B^{C_3}_{X_3}\mbox{ is equivalent to } B^{C_1}_{X_1}\union C (B^{C_2}_{X_2}\union {C'} B^{C_3}_{X_3})
\end{equation}
\begin{assumption}
All composed, constrained BNs 
$B^{C_1}_{X_1}\union C B^{C_2}_{X_2}$ used in this paper are sound.
\end{assumption}

\subsection{Non-Linear Optimization}
\label{section:nonlinearoptimization}
We next relate the judgments $\maysat$ and $\mustsat$ to optimization
problems that seek to minimize or maximize values of terms $t$ of
interest in a constrained BN $B^C_X$, and where $B^C_X$ itself may well be the result of a composition of constrained BNs as just described. We define the set of ``concretizations'' of term $t$ for $B^C_X$:
\begin{definition}
Let $t$ be a term whose variables are all in $X$ for a constrained BN
$B^C_X$. Then $\tval t\subseteq \Reals$ is defined as set $\{ \means t\alpha\mid B^C_X[\alpha]\in \concrete {B^C_X}\}$.
\end{definition}

Note that $\tval t$ does depend on $C$ and $X$ as well, but this
dependency will be clear from context. We can compute approximations
of $\sup\tval t$ and $\inf\tval t$, assuming that these values are
finite. To learn that $\sup\tval t$ is bounded above by a real $high$, we
can check whether $B^C_X\mustsat t\leq high$ holds. To learn whether
$\sup\tval t$ is bounded below by a real $low$, we can check whether
$B^C_X\maysat low\leq t$ holds. Gaining such knowledge involves both judgments
$\mustsat$ and $\maysat$. So we
cannot compute approximations of $\sup\tval t$ directly in the
existential fragment of $\mathcal Q$ but 
search for approximations by repeatedly deciding such judgments. 

We want to do this without making any assumptions about the
implementation of a decision procedure for logic $\mathcal Q$ or its
existential fragment. This can be accommodated through the use of
\emph{extended binary search}, 
as seen in Figure~\ref{fig:approxsup}, to derive an algorithm $\Sup$ 
for computing a closed interval $[low,high]$ of length at most $\delta
> 0$
such that $\sup\tval t$ is guaranteed to be in $[low,high]$.
This algorithm has as input a constrained BN $B^C_X$ with
$X$ as set of variables for constraint set $C$, a term $t$ in
$\mathcal T[X]$, and a desired accuracy $\delta > 0$. 
This algorithm assumes that $B^X_C$ is consistent and that $0 <
\sup\tval t < \infty$. We explain below how we can weaken those
assumptions to $\sup\tval t < \infty$.
\begin{figure}[ht]
$\Sup(t, \delta, B^C_X)\ \{$

\narrow
\ \ $let\ \alpha\colon X\to \Reals\ make\ \mayred{B^C_X}{t > 0}\ true;$

\narrow
\ \ $cache = \means t\alpha;$

\narrow
\ \ $while\ (\mayred{B^C_X}{t \geq 2*cache}\ satis\!f\!iable)\ \{$

\narrow
\ \ \ \ $let\ \alpha'\colon X\to \Reals\ make\ \mayred{B^C_X}{t \geq 2*cache}\ true;$

\narrow
\ \ \ \ $cache = \means t{\alpha'};$

\narrow
\ \ $\}$

\narrow
\ \ $low = cache;\ high = 2*cache;$

\narrow
\ \ $assert\  ((\mayred{B^C_X}{t\geq low}\ satis\!f\!iable)  \,\&\&\, (\mayred{B^C_X}{t\geq high}\ unsatis\!f\!iable));$

\narrow
\ \ $while\ (\mid high-low\mid {} > \delta)\ \{$

\narrow
\ \ \ \ $i\!f\ (\mayred{B^C_X}{t\geq low \, + \mid high-low\mid/{}\,2})\ satis\!f\!iable)\ \{$

\narrow
\ \ \ \ \ \ $low = low \, + \mid high-low\mid/{}\,2;$

\narrow
\ \ \ \ \ \ $assert\  ((\mayred{B^C_X}{t\geq low}\ satis\!f\!iable) \,\&\&\, (\mayred{B^C_X}{t\geq high}\ unsatis\!f\!iable));$

\narrow
\ \ \ \ $\}\ else\ \{$

\narrow
\ \ \ \ \ \ $high = low \, + \mid high-low\mid/{}\,2;$

\narrow
\ \ \ \ \ \ $assert\  ((\mayred{B^C_X}{t\geq low}\ satis\!f\!iable) \,\&\&\, \mayred{B^C_X}{t\geq high}\ unsatis\!f\!iable));$

\narrow
\ \ \ \ $\}$

\narrow
\ \ $\}$

\narrow
\ \ $return\ [low,high];$

\narrow
$\}$
\caption{Algorithm for approximating $\sup\tval t$ up to
  $\delta > 0$ for a consistent, constrained BN $B^C_X$ and term $t$
  with variables in $X$ when $0 < \sup\tval t < \infty$. The
returned closed interval $[low,high]$ has length $\leq \delta$ and
contains $\sup\tval t$. Key invariants are given as $asserts$ \label{fig:approxsup}}
\end{figure}

Algorithm $\Sup$ first uses a satisfiability witness $\alpha$ to
compute a real value $\means t\alpha$ that $t$ can attain for some
$B^C_X[\alpha]$ in $\concrete {B^C_X}$ such that $\alpha(t) > 0$.  It
then stores this real value in a $cache$ and increases the value of
$cache$ each time it can find a satisfiability witness that makes the
value of $t$ at least twice that of the current $cache$ value. Since
$\sup\tval t < \infty$, this $while$ loop terminates. The subsequent assignments to $low$ and $high$ establish an invariant that there is a value in $\tval t$ that is greater or equal to $low$, but that there is no value in $\tval t$ that is greater or equal to $high$.

The second $while$ statement maintains this invariant but makes
progress using bisection of the interval $[low,high]$. This is
achieved by deciding whether there is a value in $\tval t$ that is
greater or equal to the arithmetic mean of $low$ and $high$. If so,
that mean becomes the new value of $low$, otherwise that mean becomes
the new value of $high$. By virtue of 
these invariants, the returned
closed interval $[low,high]$ contains $\sup\tval t$ as desired. We
capture this formally:
\begin{theorem}
\label{theorem:apprsup}
Let $B^C_X$ be a consistent constrained BN and $\delta > 0$. Let
$0 < \sup\tval t < \infty$. Then we have:
\begin{enumerate}
\item Algorithm $\Sup(t, \delta, B^C_X)$
terminates, $\sup\tval t$ is in the returned closed interval
$[l,h]$ of length $\leq\delta$, and $B^C_X\maysat t\geq l$ is true.

\item Let $c$ be the initial value of $cache$.
Then the algorithm makes at most $\lfloor 2\cdot \log_2(\sup\tval t) - \log_2(c) -
  \log_2(\delta) + 1\rfloor$ satisfiability checks for formulas
  $\mayred{B^C_X}{t \geq r}$ or $\mayred{B^C_X}{t > r}$, and these
  formulas only
  differ in the choice of comparison operator and in the value of real constant $r$.
\end{enumerate}
\end{theorem}

We now give an example of using algorithm $\Sup$. Our specifications of
optimization algorithms such as that of algorithm $\Sup$ in
Figure~\ref{fig:approxsup}
do not return witness
information, we omitted such details for sake of simplicity.
\begin{example}
For constrained BN $B^{C_0}_{X_0}$ of 
Figure~\ref{fig:cBNexample},
$\Sup(mp_W, \delta,
B^{C_0}_{X_0})$ terminates for $\delta = 0.000000001$ with output
$[0.663714282364,0.663714291751]$. The value
$0.663714282364$ is attained when $x$ equals $0.299999999188$.
\end{example}

An algorithm $\Inf(t, \delta, B^C_X)$ is defined in Figure~\ref{fig:approxinf}.
It assumes that $B^C_X$ is consistent and that
$\inf\tval t$ is a subset of $\Reals_0^+$ and
contains a positive real~--~conditions we will weaken below. 
In that case, it terminates and 
returns a closed interval $[l,h]$ such that $\inf\tval t$ is in
$[l,h]$. We prove this formally:
\begin{figure}[ht]
$\Inf(t, \delta, B^C_X)\ \{$

\narrow
\ \ $let\ \alpha\colon X\to \Reals\ make\ \mayred{B^C_X}{t > 0}\ true;$

\narrow
\ \ $cache = \means t\alpha;$

\narrow
\ \ $while\ (\mayred{B^C_X}{t \leq 0.5*cache}\ satis\!f\!iable\ and\
0.5*cache > \delta)\ \{$

\narrow
\ \ \ \ $let\ \alpha'\colon X\to \Reals\ make\ \mayred{B^C_X}{t \leq 0.5*cache}\ true;$

\narrow
\ \ \ \ $cache = \means t{\alpha'};$

\narrow
\ \ $\}$

\narrow
\ \ $i\!f\ (\mayred{B^C_X}{t \leq 0.5*cache}\ satis\!f\!iable)\ \{\ return\ [0,0.5*cache];\ \}$

\narrow
\ \ $low = 0.5*cache;\ high = cache;$

\narrow
\ \ $assert\  (\mayred{B^C_X}{t\leq low}\ unsatis\!f\!iable) \,\&\&\, (\mayred{B^C_X}{t\leq high}\ satis\!f\!iable));$

\narrow
\ \ $while\ (\mid high-low\mid {} > \delta)\ \{$

\narrow
\ \ \ \ $i\!f\ (\mayred{B^C_X}{t\leq low \, + \mid high-low\mid/{}\,2})\ satis\!f\!iable)\ \{$

\narrow
\ \ \ \ \ \ $high = low \, + \mid high-low\mid/{}\,2;$

\narrow
\ \ \ \ \ \ $assert\  ((\mayred{B^C_X}{t\leq low}\ unsatis\!f\!iable) \,\&\&\, (\mayred{B^C_X}{t\leq high}\ satis\!f\!iable));$

\narrow
\ \ \ \ $\}\ else\ \{$

\narrow
\ \ \ \ \ \ $low = low \, + \mid high-low\mid/{}\,2;$

\narrow
\ \ \ \ \ \ $assert\  ((\mayred{B^C_X}{t\leq low}\ unsatis\!f\!iable) \,\&\&\, (\mayred{B^C_X}{t\leq high}\ satis\!f\!iable));$

\narrow
\ \ \ \ $\}$

\narrow
\ \ $\}$

\narrow
\ \ $return\ [low,high];$

\narrow
$\}$
\caption{Algorithm for approximating $\inf\tval t$ up to
  $\delta > 0$ for a consistent, constrained BN $B^C_X$ and term $t$
  in $\mathcal T[X]$ when $\tval t\subseteq \Reals_0^+$ contains a positive real.
\label{fig:approxinf}}
\end{figure}

\begin{theorem}
\label{theorem:apprinf}
Let $B^C_X$ be a consistent constrained BN and $\delta > 0$. Let
$\tval t\subseteq \Reals_0^+$ contain a positive real.
Then we have:
\begin{enumerate}
\item Algorithm $\Inf(t, \delta, B^C_X)$
terminates and $\inf\tval t$ is in the returned interval
$[l,h]$ such that $h-l\leq\delta$ and $B^C_X\maysat t\leq h$ are true.

\item Let $c$ be the initial value of $cache$. Then the algorithm
  makes one
  satisfiability check $\mayred{B^C_X}{t > 0}$ and at
  most  $\lfloor 2\cdot \log_2(c) - \log_2(\min(\delta,\inf\tval t))\rfloor$
  satisfiability checks for formulas $\mayred{B^C_X}{t \leq r}$, and
  these formulas only
  differ in the size of real constant $r$.
\end{enumerate}
\end{theorem}

We now show how we can relax the conditions of $B^C_X$ being consistent
and of $0 < \sup\tval
t < \infty$ to $\sup\tval t < \infty$. 
In Figure~\ref{fig:supwrapper}, we see this modified algorithm
$\Sup^*$ which relies on both $\Sup$ and $\Inf$. It returns a
closed interval with the same properties as that returned by
$\Sup$ but where $\sup\tval t$ only need be finite. 
We state the correctness of this algorithm formally:
\begin{figure}
$\Sup{}^{\wrap}(t, \delta, B^C_X)\ \{$

\narrow
\ \ $i\!f\ (\mayred{B^C_X}{t > 0}\ satis\!f\!iable)\ \{$

\narrow
\ \ \ \ $return\ \Sup(t,\delta,B^C_X);\ \}$

\narrow
\ \ $elsei\!f\ (\mayred{B^C_X}{t = 0}\ satis\!f\!iable)\ \{$

\narrow
\ \ \ \ $return\ 0\ as\ maximum\ f\!or\ t;\ \}$

\narrow
\ \ $elsei\!f\ (\mayred{B^C_X}{t < 0}\ satis\!f\!iable)\ \{$

\narrow
\ \ \ \ $let\ [l,h] = \Inf(-t,\delta,B^C_X);$

\narrow
\ \ \ \ $return\ [-h,-l];$

\narrow
\ \ $\}$

\narrow
\ \ $return\ B^C_X\ is\ inconsistent;$

\narrow
$\}$
\caption{Algorithm $\Sup{}^{\wrap}$ uses algorithms $\Sup$ and $\Inf$ and
terminates whenever $\sup\tval t < \infty$. It either recognizing that $0$ is the maximum of $\tval
  t$, returns a closed interval $[l,h]$ with $h-l\leq \delta$
  such that $\sup\tval t$ is in $[l,h]$, or it detects that $B^C_X$
  is inconsistent
\label{fig:supwrapper}}
\end{figure}

\begin{theorem}
\label{theorem:supwrapper}
Let $B^C_X$ be a constrained BN,
$\delta > 0$, and $\sup\tval t < \infty$. Then
  $\Sup{}^{\wrap}(t,\delta,B^C_X)$ terminates and its calls to
  $\Sup$ and $\Inf$ meet their preconditions. Moreover, it either correctly identifies that $B^C_X$ is inconsistent, that $0$ is the maximum of $\tval t$ or it returns a closed
  interval $[l,h]$ such that $\sup\tval t$ is in that interval, $h-l\leq$ is less than or equal to $\delta$, and $B_X^C\maysat t\geq l$ holds.
\end{theorem}

We conclude this section by leveraging $\Sup{}^{\wrap}$ to an
algorithm $\Inf{}^{\wrap}$, seen in Figure~\ref{fig:infwrapper}. Algorithm $\Inf{}^{\wrap}$
relaxes that $\tval t$ contains a
positive real and is a subset of $\Reals_0^+$ to a more general
pre-condition $-\infty < \inf\tval
t$, and it has correct output for inconsistent, constrained BNs. We formalize this:
\begin{figure}
$\Inf{}^{\wrap}(t, \delta, B^C_X)\ \{$

\narrow
\ \ $let\ x = \Sup{}^{\wrap}(-t, \delta, B^C_X);$

\narrow
\ \ $i\!f\ (x\ reports\ that\ B^C_X\ is\ inconsistent)\ \{\ return\ B^C_X\ is\ inconsistent;\ \}$

\narrow
\ \ $elsei\!f\ (x\ reports\ 0\ as\ maximum\ f\!or\ -\!t)\ \{\ return\ 0\ as\ minimum\ f\!or\ t;\ \}$

\narrow
\ \ $elsei\!f\ (x\ reports\ interval\ [l,h])\ \{\ return\ [-h,-l]; \ \}$

\narrow
$\}$
\caption{Algorithm $\Inf{}^{\wrap}$ uses algorithm $\Sup{}^{\wrap}$ and terminates whenever $-\infty <
  \inf\tval t$. It either recognizes that $0$ is the minimum of $\tval
  t$, returns a closed interval $[l,h]$ with $h-l\leq \delta$
  such that $\inf\tval t$ is in $[l,h]$, or it detects that $B^C_X$
  is inconsistent
\label{fig:infwrapper}}
\end{figure}

\begin{theorem}
\label{theorem:infimum}
Let $B^C_X$ be a constrained BN, $\delta > 0$, and $t$ a term with $-\infty < \inf \tval t$. Then $\Inf{}^{\wrap}(t,\delta,B^C_X)$ terminates and either correctly identifies that $B^C_X$ is inconsistent, that $0$ is the minimum of $\tval t$ or it returns a closed interval $[l,h]$ of length $\leq \delta$ such that $\inf\tval t$ is in $[l,h]$ and $B_X^C\maysat t\leq h$ holds.
\end{theorem}

Let us revisit Example~\ref{examplediff} to illustrate use of $\Sup{}^{\wrap}$.
\begin{example}
Let $\tilde C_0$ be $C_0\cup \{0.1 \leq x \leq 0.2\}$. For constrained BN $B^{\tilde
  C_0}_{X_0}\union C B^{C'_0}_{X'_0}$, we maximise $\diff$ using
$\Sup{}^{\wrap}(\diff, 0.000000001, B^{\tilde C_0}_{X_0}\union C
B^{C'_0}_{X'_0})$, which returns the interval
\[
[-0.055219501217,-0.0552194960809]
\]

\noindent arising from the
third case of $\Sup{}^{\wrap}$ as both $\mayred{B^C_X}{t > 0}$ and
$\mayred{B^C_X}{t = 0}$ are unsatisfiable, but formula $\mayred{B^C_X}{t < 0}$
is satisfiable. It shows that marginal $mp_W$ is always smaller
than marginal $mp'_W$ in this constrained BN, in contrast to the situation of Example~\ref{examplediff}.
\end{example}

\section{Detailed Case Study}
\label{section:casestudy}
We now apply and evaluate the foundations for
constrained BNs on a case study in the context of arms control. 
Article VI of the Treaty on the Non-Proliferation of
Nuclear Weapons (NPT) \cite{npt70} states that each treaty party 
\begin{quote}
\emph{``undertakes to pursue negotiations in good faith on effective measures relating to cessation of the nuclear arms race at an early date and to nuclear disarmament, and on a treaty on general and complete disarmament under strict and effective international control.''}
\end{quote}

\noindent One important aspect of meeting such treaty obligations may
be the creation and execution of trustworthy inspection processes, for
example to verify that a treaty-accountable item has been made
inoperable. Designing such processes 
is challenging as it needs to guarantee sufficient mutual
trust between the inspected and inspecting party in the presence of
potentially conflicting interests. Without such trust, the parties might 
not agree to conduct such inspections.

The potential benefit of mathematical models for the design and evaluation of such
inspection processes is apparent. Bayesian Networks can capture a form
of trust~--~through an inherent bias of processing imperfect information~--~
and different degrees of beliefs~--~expressed, e.g., in subjective
probabilities. Bayesian Networks can also represent objective data
accurately, and their graphical formalism may be understood
by domain experts such as diplomats. These are good reasons for
exploring Bayesian Networks for modeling and evaluating
inspection processes. But Bayesian Networks do not seem to have means of
\emph{building confidence} in their adequacy and utility, especially
in this domain in which prior data for 
learning both graph structure and probabilities at nodes in such a
graph are hard to find. We now show how \emph{constrained} BNs can be
used to build such confidence in mathematical models of an inspection process.

\subsection{An Arms Control Inspection Process}
Consider the situation of two fictitious nation states.
The \emph{inspecting} nation is tasked with identifying whether an item belonging to the \emph{host} nation, available to inspect in a controlled inspection facility and declared by the host nation to be a nuclear weapon,
is indeed a nuclear weapon. This situation is similar to a scenario that had
been explored in the UK/Norway initiative in 2007 \cite{UKNIwebsite, norway10}.

Given the nations' non-proliferation obligations and national security
concerns, the design details of the inspected item must be
protected: the inspecting nation will have no visual
access to the item.  Instead the nations agree that the
to-be-inspected item contain Plutonium with the isotopic ratio
240Pu:239Pu below a certain threshold value, which they set at 0.1.

\begin{figure}
\centering
\includegraphics[scale=0.207]{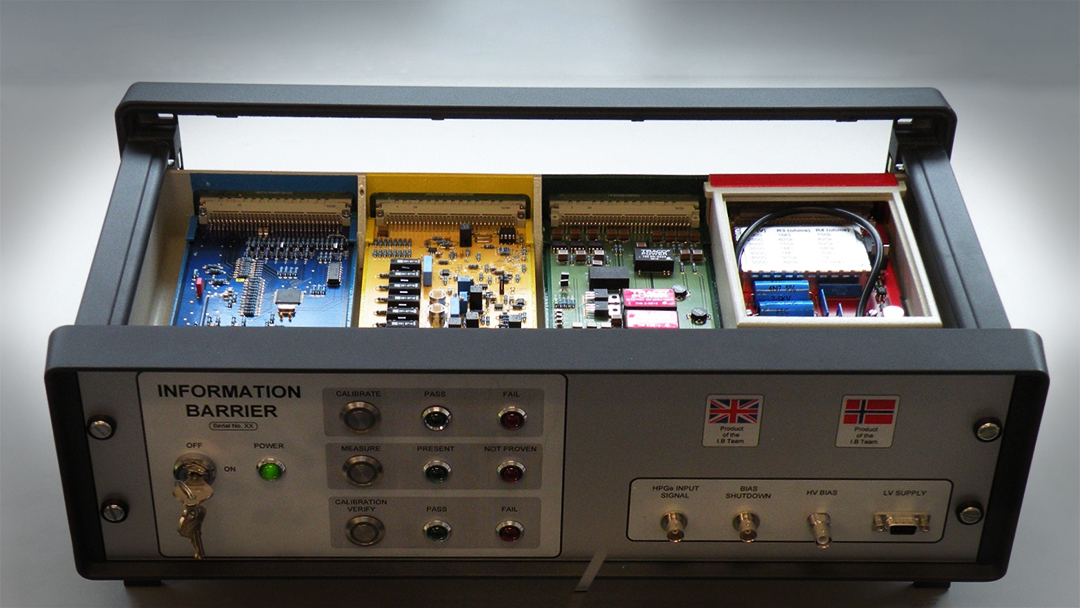}
\caption{Prototype of an information barrier (IB), photo taken from
  \cite{UKNIwebsite} ,  that the inspecting nation might
  build. The IB would output either a green or red light to inspectors based on
  physical measurements made within the IB that verify the presence of nuclear material\label{fig:IB}}
\end{figure}

In order to draw conclusions about whether an item presented for
inspection is a weapon, the inspecting nation uses an information
barrier (IB) system comprising a HPGe detector and bespoke electronics
with well-understood performance characteristics (see Figure
\ref{fig:IB}, \cite{UKNIwebsite}) to conduct measurements on the
item while the item is concealed in a box. The IB system displays
a green light if it detects a gamma-ray spectrum indicative of the
presence of Plutonium with the appropriate isotopic ratio; if it does
not detect this spectrum for whatever reason, it shows a red
light. No other information is provided, and weapon-design information
is thus protected \cite{norway10}.

The inspecting nation believes that it may be possible for the host nation
to spoof a radioactive signal~--~~or in some way provide a surrogate~--~to
fool the detector, or that the host nation may have just placed Plutonium
with the appropriate isotopic ratio in the box rather than a weapon.
These subjective assessments should be reflected in the mathematical model
alongside the error rates of the IB system that reflect the
reliability of that device. 

In order to deter cheating, the inspecting nation is allowed to choose the
IBs used in the verification from a pool of such devices provided by
the host nation, and may choose one or two IBs to that end. 
From that same pool of devices, the inspecting nation may take some unused IBs away for
authentication~--~activities designed to assess whether the host
nation tampered with the IBs. But the inspecting nation must not inspect any
\emph{used} IBs, to prevent it from exploiting any residual
information still present in such used IBs to its advantage. 

This selection process of IBs is therefore designed to ensure that a
nefarious host nation is held at risk of detection should it decide to tamper
with the IBs used in verification: it would run the risk of one or
more tampered IBs being selected for authentication by the inspecting
nation. Although such authentication cannot be assumed to be perfect~--~and
 this fact, too, should be modelled~--~the prospect of detection may
deter such a host. 

\begin{figure}
\begin{center}
\includegraphics[angle=90,scale=0.35	]{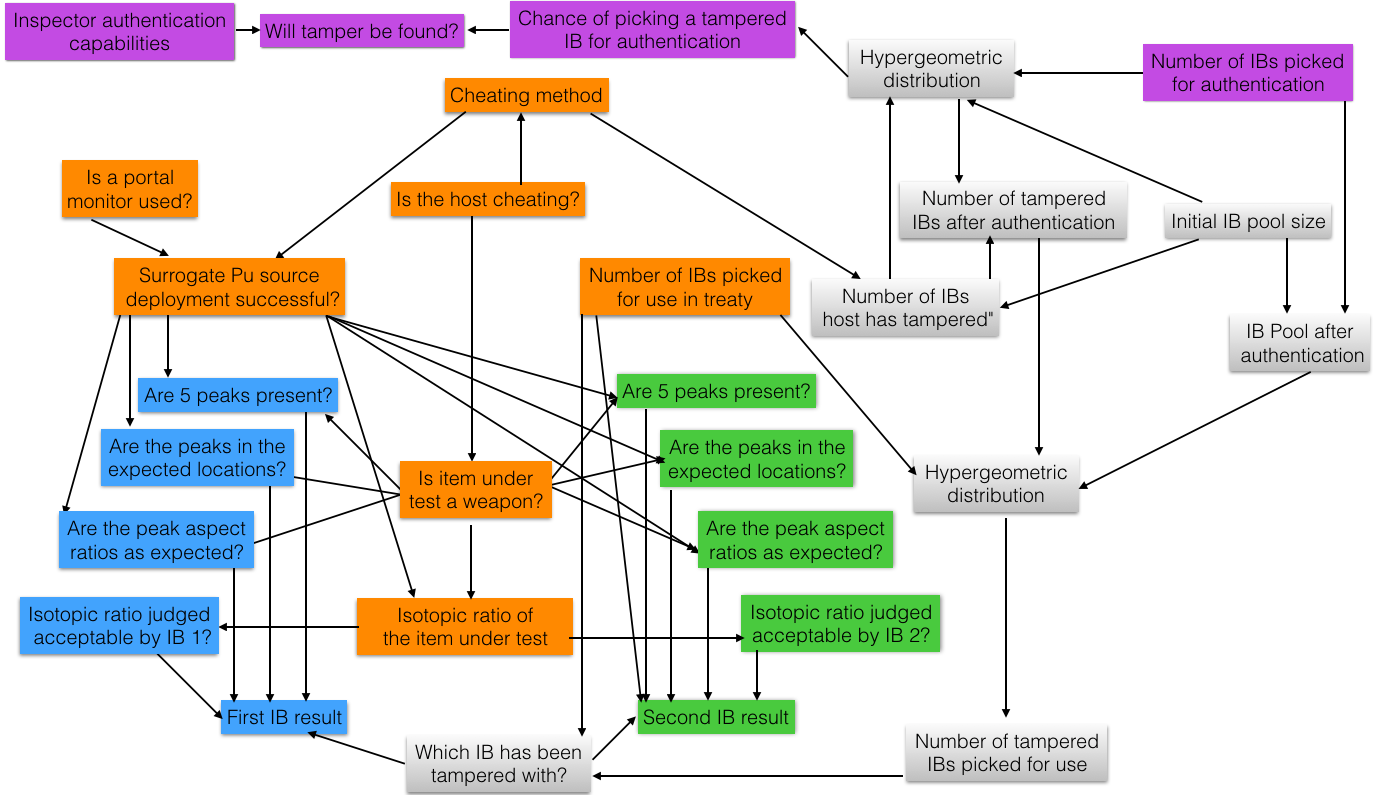}
\end{center}
\caption{A BN \cite{INMMpaper} which details
  aspects of an arms inspection process. Aspects of the 1-2 IB devices used for verification
are modeled in blue and 
  green nodes, respectively. The assessment of cheating and the
  operation of the inspection in other ways are shown in orange nodes,
  and authentication procedures are modeled in purple
  nodes. Mathematical or logical computations 
are represented in grey nodes. \label{fig:inmmmodel}}
\end{figure}

We model this inspection process through constrained BNs that
are abstracted from a sole BN with DAG 
shown in Figure~\ref{fig:inmmmodel} and based on a
design developed by the Arms Control Verification Research group at AWE. 
This DAG depicts different aspects of the verification procedure in four key
areas: 
\begin{itemize}
\item the \emph{selection of the IBs} for inspection or 
authentication purposes, 

\item the \emph{workings of the IB} in the
inspection itself, 

\item \emph{authentication} of (other) IBs, and 

\item the combination of these aspects to assess
\emph{any possibility of cheating overall}, be it through IB tampering, surrogate nuclear sources, and so forth. 
\end{itemize}

The selection of the IBs starts with the {\sf IB pool size}; a
selection of IBs built by the host nation, from which there will be
a {\sf Number of IBs picked for authentication} and {\sf Number of IBs
  picked for use} by the inspecting nation. Should a {\sf
  Number of tampered IBs} exist, then the selection process (blind to
such a tamper) follows a {\sf Hypergeometric distribution} and will
probabilistically determine whether such tampered IBs make it into use
in the verification process, authentication process or neither. The
choice of distribution reflects that IBs~--~once chosen for either verification or authentication~--~cannot be used for any other purpose.

The IBs picked for either authentication or
verification help the inspecting nation to judge whether the item
under inspection {\sf Is a weapon}. A weapon or a {\sf Surrogate Pu
  source} determine physical nuclear properties about the {\sf
  Isotopic ratio} of Plutonium elements.  Our mathematical model
captures a possible inspector judgment that a surrogate source would
only be used if the host felt that it was extremely likely to pass the
IBs verification tests. Therefore,
any surrogate source would have isotopic properties at least as good as those of a real weapon. 

We stress that the probabilities chosen for each isotopic ratio,
conditioned on whether the item under test is or is not a weapon, are \emph{not} derived from real-world weapons data, but instead reflect in broad terms that Plutonium with a higher isotopic ratio than the chosen threshold is less likely to be found in a nuclear weapon. A bespoke algorithm is used by the IB system on the collected gamma-ray spectrum to test whether both the {\sf Peaks are in the expected locations} and the {\sf Peak aspect ratio} are as expected. If all {\sf 5 peaks are present} and the {\sf Ratio of 240/239 isotopes is acceptable}, then one or both of the {\sf First IB result} or {\sf Second IB result} are reported, conditional on any tampering and depending on whether or not two IBs are used to test the same item.

A mathematical model cannot hope to reflect each potential  tamper. Therefore, we model
authentication as an assessment of the {\sf Inspector's authentication
  capabilities}: the better these are, the more likely the
{\sf Tamper will be found}, and this requires that
at least one tampered IB exists and was selected for authentication.
This is
controlled by the parent nodes: the aforementioned {\sf Hypergeometric distribution}, and a node {\sf Chance of picking a tampered IB for authentication}.

The mathematical model is drawn together by the overarching question of ``Is the
{\sf Host cheating?}''. If so, we then determine a {\sf Cheating
  method}, which reflects the understanding of the inspecting nation about the
possible ways that the host nation could try to cheat, as outlined above, and
the prior beliefs of the inspecting nation about the relative
likelihood of the use of each method
if the host nation were to be cheating.

Finally, we check whether a {\sf Portal monitor is used} to stop transportation of radioactive material~--~which could be used as a surrogate source~--~in and out of the facility, although we do not model this aspect in greater detail.

The probabilities used in this BN
stem from a variety of sources. Some are somewhat arbitrarily selected, as
described above, and therefore need means of building confidence in
their choice. Probabilities relating to the performance of the IB
system are derived from experimental analysis of the UKNI IB
\cite{norway10, UKNIwebsite}. 

The size of the probability tables for nodes
of the BN in Figure~\ref{fig:inmmmodel} range from small
(a few or tens of entries), to medium (hundreds of
entries) and larger ones (thousands of entries). Given that
complexity, we refrain from specifying more details on these tables
within the paper itself. 

Our evaluation of the methods developed in
Section~\ref{section:theory} will abstract the BN described above (see
Figure~\ref{fig:inmmmodel}) 
into constrained BNs, and demonstrate that these abstractions 
can inform decision support given the sparsity or lack of
prior data that informed its choices of probabilities. 
\begin{assumption}
For convenience, this case study will not explicitly list or show the constraints that
define the meaning of marginals symbolically. These meanings are
included in the open-access research code cited on
page~\pageref{page:URL}.
\end{assumption}

\subsection{Impact of Cheating Method on Tamper Detection}
We want to understand how the choice of cheating method can impact the
probability of detecting a tamper. The uncertainty about what cheating
method the host nation will adopt is modelled in a constrained BN
$B^{C_1}_{X_1}$ that takes the BN from Figure~\ref{fig:inmmmodel} and
replaces the probability table for its node {\sf Cheating Method} as
specified in Figure~\ref{fig:cBNone}. We use variables $x$, $y$, and
$u$ to denote, respectively, the probability of IB tamper only,
Surrogate source tamper only, and both IB tamper and surrogate source
tamper. The variable $mp_{t\!f}$ refers to the marginal probability
$p(\textnormal{{\sf Will tamper be found?} = Yes})$.
\begin{figure}
\centering
\begin{tabular}{@{}lcc@{}}
\toprule
\multicolumn{3}{c}{{\sf Cheating Method}}                           \\ \midrule
                              & Is cheating & Is not cheating \\
None                          & 0           & 1               \\
IB tamper only                & $x$         & 0               \\
Surrogate source only         & $y$         & 0               \\
IB tamper \& surrogate source & $u$         & 0               \\ \bottomrule
\end{tabular}
\caption{Probability table for node {\sf Cheating Method} in
  constrained BN $B^{C_1}_{X_1}$ where $C_1$ contains $\{ 0 < x, y, u
  < 1.0, x + y + u = 1.0   \}$, $\varx = \{x,y,u\}$, $\varmp =
  \{mp_{t\!f}\}$, $X_1 = \varx\cup \varmp$, and the BN graph  and all
  other probability tables for $B^{C_1}_{X_1}$ are as for the BN in Figure~\ref{fig:inmmmodel}\label{fig:cBNone}}
\end{figure}

We compute  the
interval $[l,h] = [0.197137598548,0.197137608314]$ as output of the
function call $\Sup{}^{\wrap}(mp_{t\!f}, 0.00000001, B^{C_1}_{X_1})$. The witness
information for the existentially quantified variables $x$, $y$, $u,$
and $mp_{t\!f}$ pertains to value $l=0.197137598548$:
\[
x = 0.000000010001153\ \ \ y = 0.000000010001153\ \ \ u = 0.999999979997693
\]

\noindent We compute  the
interval $[\tilde l, \tilde h] =
[5.875158e\!\!-\!\!09,1.1750316e\!\!-\!\!08]$ as output of function
call $\Inf{}^{\wrap}(mp_{t\!f}, 0.00000001, B^{C_1}_{X_1})$. The witness
information is now for the value $\tilde h=1.1750316e\!\!-\!\!08$ of $mp_{t\!f}$ and we get 
\[
x =  0.000000030265893\ \ \
y =  0.999999939468212\ \ \
u =  0.000000030265893\ \ \
\]

\noindent We may combine this information, for example to bound
the range of values that $mp_{t\!f}$ can possibly attain, as the 
interval
\begin{equation}
[\tilde l, h] = [0.000000000587,0.197137608314] \nonumber
\end{equation}

\noindent We therefore conclude that this marginal
probability can only vary by less than $0.19714$ in the
given strict uncertainty of the model.

Let us now ask for what values of $x$ can $mp_{t\!f}$ be within $0.01$
of the lower bound $l= 0.197137598548$ returned for $\Sup{}^{\wrap}$
above. To that end, we consider the constrained BN $B^{C_1'}_{X_1}$
where
\(C_1' = C_1\cup \{ \mid mp_{t\!f} - 0.197137598548 \mid {}\leq{} 0.01 \}\)
and compute lower and upper bounds for $x$ in this
constrained BN:
\[
\begin{array}{ll}
[l^x,h^x] = \Sup{}^{\wrap}(x, 0.00000001, B^{C_1'}_{X_1}) =[0.999999994824,1.00000000196]& {} \\
\end{array}
\]
\[
\begin{array}{ll}
[ {\tilde l}^x,{\tilde h}^x ] = \Inf{}^{\wrap}(x, 0.00000001, B^{C_1'}_{X_1}) = [7.4505805e\!\!-\!\!09,1.4901161e\!\!-\!\!08]& {}
\end{array}
\]

\noindent From this we can learn that 
\begin{eqnarray}
\forall x\colon \bigl [ (1.4901161e\!\!-\!\!08 \leq x\leq 0.999999994824)\land
\bigwedge C_1\bigr ] \rightarrow &{}& {}\nonumber\\
\ \mid mp_{t\!f} - 0.197137598548 \mid {}\leq{} 0.01 &{}& {} \label{equ:xwithin}
\end{eqnarray}

\noindent is logically valid: whenever $x$ is in that value range and
all constraints in $C_1$ are satisfied (which is true for all
concretizations of $B^{C_1}_{X_1}$), then the marginal $mp_{t\!f}$ is
within $0.01$ of the lower bound for its maximal value.
 
 Repeating these optimizations above for variables $y$ and $u$, we
 determine similar formulas that are logically valid:
\begin{eqnarray}
\forall y\colon \bigl [ (1.209402e\!\!-\!\!08 \leq y \leq 0.0507259986533)\land
\bigwedge C_1\bigr ] \rightarrow &{}& {}\nonumber\\
\ \mid mp_{t\!f} - 0.197137598548 \mid {}\leq{} 0.01 &{}& {}
\nonumber\\
\forall u\colon \bigl [ (1.4901161e\!\!-\!\!08 \leq u \leq 0.999999998164)\land
\bigwedge C_1\bigr ] \rightarrow &{}& {}\nonumber\\
\ \mid mp_{t\!f} - 0.197137598548 \mid {}\leq{} 0.01 &{}& {} \nonumber
\end{eqnarray}

These results say that the marginal $mp_{t\!f}$ is insensitive
to changes to $x$, which is able to vary across the whole range $(0.0,1.0)$
without having much impact on the results; the situation is very
similar for variable $u$. For variable $y$, the range at which $mp_{t\!f}$ is
not too sensitive on changes of $y$ is much smaller~--~just over
$0.05$. Overall, we conclude that the model remains
in the area of highest probability for detecting tampering as long
as $x$ or $u$ are large.

Our analysis shows that
the ``tamper'' cheating method is the one for which there is the
highest chance of detecting cheating. However, our results also
highlight that unless both tamper and surrogate source, or tamper on
its own are used, there are limited ways in which to detect cheating
through these nodes.  From this we learn that use of a portal monitor
is advisable, as any increase in $y$ moves the marginal out of the
region of highest probability of detecting cheating, and decreases the
chance of cheating being detected otherwise.  
Related to this is that the range of $y$ gives potential insight into
future work required on tamper detection for the inspecting
nation. Despite contributing neither to an IB tamper nor detection, $y$ can
vary by over 0.05~--~over five times that of the movement away from the
marginal $mp_{t\!f}$'s maximum point by only 0.01. This suggests there
are other limiting factors to tamper detection, such as
capability, that could be better reflected in a mathematical model.

\subsection{Comparing two BN models}
We now illustrate the benefits of composing two constrained BNs (see
Section~\ref{subsection:union}). Two constrained BNs, $B^{C_2}_{X_2}$
and $B^{C'_2}_{X'_2}$, are defined in Figure~\ref{fig:cBNfour}. Both
have symbolic and equivalent probability tables for node {\sf
  Authentication Capabilities} but consider different hard evidence for the probability of a
tamper to be found. In $B^{C_2}_{X_2}$,
there is $1$ IB machine picked for authentication whereas in $B^{C'_2}_{X'_2}$
there are $5$ IB machines picked to that end, resulting in the respective marginals
\begin{eqnarray}
p(\textnormal{{\sf Will tamper be found?} = Yes} \mid  \textnormal{{\sf
    Host cheating} = Yes,} \\ \nonumber \textnormal{{\sf Number of IBs picked for authentication} = 1}) {} &{}&\ \  \label{equ:marginalCfour}\\
p(\textnormal{{\sf Will tamper be found?} = Yes} \ | \ \textnormal{{\sf
    Host cheating} = Yes,} \\ \nonumber \textnormal{{\sf Number of IBs picked for authentication} = 5}) {} &{}&\ \  \label{equ:marginalCfourprime}
\end{eqnarray}
\begin{figure}
\centering
\begin{tabular}{@{}lcc@{}}
\toprule
\multicolumn{3}{c}{{\sf Authentication Capabilities}}                           \\ \midrule
                          Good & Medium & Poor \\
$x$ & 0.3333 & $0.6667-x$           \\ \bottomrule
\end{tabular}
\caption{
Probability table for node  {\sf Authentication Capabilities}  in
constrained BN $B^{C_2}_{X_2}$ that is like that BN in
Figure~\ref{fig:inmmmodel} except that the symbolic probability table
for node {\sf Authentication Capabilities} is as above, $\varx =
\{x\}$, $\varmp = \{mp_{t\!f_2}\}$, and $C_2$ contains $\{0\leq x\leq
0.6667, x + 0.3333 + (0.6667-x) = 1\}$. Variable $mp_{t\!f_2}$ denotes
the marginal in~(\ref{equ:marginalCfour}). Constrained BN
$B^{C'_2}_{X'_2}$ is a ``copy'' of $B^{C_2}_{X_2}$ that replaces all
occurrences of $x$ with $y$ 
 and has $\varmp = \{mp_{t\!f'_2}\}$ where 
$mp_{t\!f'_2}$ denotes the marginal in~(\ref{equ:marginalCfourprime})\label{fig:cBNfour}}
\end{figure}

\noindent In both models, the probability for state ``Good'' is 
bounded by $0.6667$ so that there is
a ``gradient'' pivoting around {\sf Medium} capabilities fixed at 0.3333. 

We seek decision support on how much to prioritise research into IB authentication capabilities, each of $B^{C_2}_{X_2}$ and $B^{C'_2}_{X'_2}$ representing a different capability scenario. Of interest here is the change in the likelihood that a tamper will be found.
We can simply model this by defining a new term
\begin{equation}
\diff = mp_{t\!f_2} - mp_{t\!f'_2}
\label{equ:difffour}
\end{equation}

\begin{figure}
\begin{center}
\includegraphics[scale=0.3]{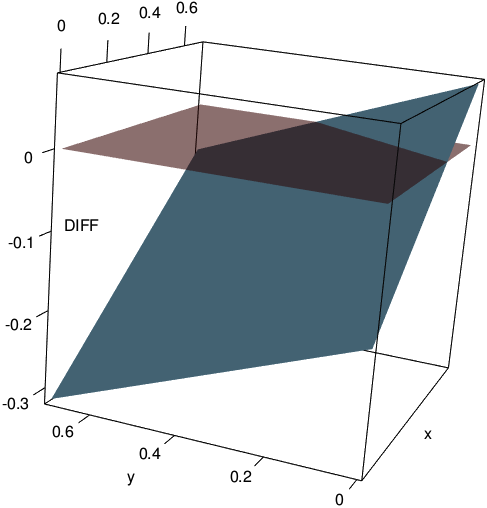}
\end{center}
\caption{The surface in blue shows the values of $\diff$
  in~(\ref{equ:difffour}) for the sample points
  defined in~(\ref{equ:samplepoints}). The plane $\diff = 0$ is shown
  in red. Its intersection with the blue surface marks the boundary of
  where decision support would favor running $1$ IB authentication (above the
  red plane) and $5$ IB authentications (below the red plane). The linear equation
  defining this boundary line is 
  \(0= -0.212884507399337\!*\! x +0.354426098468987\!*\! y -0.16515544651\).
\label{fig:system1}}
\end{figure}

\noindent Variable $\diff$ is in $\varx$ for the constrained BN $B^{C_2}_{X_2}\union C B^{C'_2}_{X'_2}$ where the constraint set $C$ for this combination is $\{\diff = mp_{t\!f_2} - mp_{t\!f'_2}\}$.
We compute the value of $\diff$ for each combination of values $(x,y)$ from set
\begin{equation}
S = \{(0.0+0.01\cdot a,0.0+0.01\cdot b)\mid 0\leq a,b\leq 67\}
\label{equ:samplepoints}
\end{equation}

\noindent and linearly interpolate 
the result as a surface seen in~Figure \ref{fig:system1}. The
linear relationship between the symbolic probabilities of node
{\sf Authentication Capability} to that of its child node {\sf Will
  the tamper be found?} make this surface flat.

We can now use the method familiar from our earlier analyses to assess the value range of term $\diff$ in this composed, constrained BN.
The function call $\Sup{}^{\wrap}(\diff, 0.00000001, B^{C_2}_{X_2}\union C B^{C'_2}_{X'_2})$ returns the
interval
\begin{equation}
[l,h] = [0.0711404333363,0.0711404338663] \nonumber
\end{equation}

Next,
$\Inf{}^{\wrap}(\diff, 0.00000001, B^{C_2}_{X_2}\union C B^{C'_2}_{X'_2})$ is computed as the
interval
\[
[\tilde l, \tilde h] = [-0.307085548061,-0.307085547533]
\]

\noindent In particular, the values of
$\diff$ for all concretizations of $B^{C_2}_{X_2}\union C B^{C'_2}_{X'_2}$
lie in the interval
\begin{equation}
[-0.307085548061, 0.0711404338663] \nonumber
\end{equation}

\noindent The blue surface of $\diff$ in Figure~\ref{fig:system1} is mostly negative (below the red plane).
This shows that the case of testing $5$ IBs for tampers is nearly always better, irrespective of the confidence one may have in one's ability to find a tamper. This is true, other than for the most extreme cases when there is the least confidence in authentication capabilities when testing five IBs (for $y = 0$) and most confidence when testing one (for $x = 0.667$).

Let us next explore a situation in which the inspector believes to have high authentication capabilities, regardless of whether $1$ or $5$ IBs are picked for authentication. We can easily model this by setting
$C' = C\cup \{0.467\leq x,y\leq 0.667\}$ and refining the composed
model using $C'$. We compute the output of
$\Sup{}^{\wrap}(\diff, 0.00000001, B^{C_2}_{X_2}\union C' B^{C'_2}_{X'_2})$ to be the
interval
\begin{equation}
[l,h] = [-0.0282766319763,-0.0282766314489] \nonumber
\end{equation}

\noindent and $\Inf{}^{\wrap}(\diff, 0.00000001, B^{C_2}_{X_2}\union C' B^{C'_2}_{X'_2})$ to be the
interval
\begin{equation}
[\tilde l, \tilde h] = [-0.141568560141,-0.141568559299] \nonumber
\end{equation}

\noindent Now $\diff$ is in $[-0.141568560141,-0.0282766314489]$ and the
largest absolute difference between picking $1$ and $5$ IBs for
authentication is greater than $0.14$,
witnessed when the inspector has a particularly high capability in
authenticating $5$ IBs, (when
$y = 0.667$ and $mp_{t\!f'_2} = 0.24932$) compared with only
inspecting one IB with more moderate capability (when $x = 0.467, mp_{{t\!f}_2} = 0.10696$).

A decision maker could vary the use of the above approach in order to
weigh the cost of IB production against the cost of developing and employing more advanced authentication capabilities.  He or she could also query in detail how the results of such cost-benefit analyses might change as new information is learned or new techniques deployed. This capability might help decision makers to balance their priorities and to gain the best assurance possible within a cost budget that the verification regime they implement is effective.

\subsection{Determining equivalent decision support}
We assess the consistency of two different constrained BNs 
of equal intent of decision support.
Constrained BNs $B^{C_3}_{X_3}$ and $B^{C'_3}_{X'_3}$ are identical to
$B^{C_1}_{X_1}$ and its symbolic probability table for node {\sf
  Cheating Method} as in Figure \ref{fig:cBNone}, except that $th$ is
an additional variable 
used to model decision support.
Variable set $\varmp$ also changes. For $B^{C_3}_{X_3}$ we have $\varmp = \{mp_{t\!f_3}\}$ and for  $B^{C'_3}_{X'_3}$ we set
$\varmp = \{mp_{t\!f'_3}\}$ instead. Variable $mp_{t\!f_3}$ denotes marginal probabilities for 
hard evidence that {\sf Initial Pool Size} = 10 IBs
in~(\ref{equ:mptffive}), whereas variable $mp_{t\!f'_3}$ denotes a
marginal for hard evidence {\sf Initial Pool Size} = 20 IBs in~(\ref{equ:mptffiveprime}):
\begin{eqnarray}
p(\textnormal{{\sf Will tamper be found?} = Yes} \mid \textnormal{{\sf Initial Pool Size} = 10}) \label{equ:mptffive} \\
p(\textnormal{{\sf Will tamper be found?} = Yes} \mid \textnormal{{\sf Initial Pool Size} = 20}) \label{equ:mptffiveprime}
\end{eqnarray}

\noindent These are marginal probabilities that the nation which is authenticating IBs will find a tamper.
A decision~--~for example that an IB has been tampered with~--~may
then be supported if such a marginal is above a certain threshold $th$. We now want to understand whether the two constrained BNs would support decisions in the same manner, and for what values or value ranges of $th$. 

For any value $th$, consider the constraint $\varphi_{th}$ in $\mathcal Q$ given by
\begin{equation}
\lnot \bigl [ \bigl ( (th < mp_{t\!f_3})\land
(mp_{t\!f'_3} \leq th) \bigr )\lor \bigl ( (th < mp_{t\!f'_3})\land
(mp_{t\!f_3}\leq th) \bigr ) \bigr ] \nonumber
\end{equation}

\noindent We can now analyze whether both constrained BNs will always support decisions through threshold $th$ by evaluating
\begin{equation}
B^{C_3}_{X_3}\union C B^{C'_3}_{X'_3}\mustsat \varphi_{th}
\label{equ:samedecision}
\end{equation}

\noindent where $C$ equals $\{0 < th < 1\}$. By Theorem~\ref{theorem:dual}, judgment~(\ref{equ:samedecision}) is equivalent to
\begin{equation}
\hbox{not } B^{C_3}_{X_3}\union C B^{C'_3}_{X'_3}\maysat \bigl ( (th < mp_{t\!f_3})\land
(mp_{t\!f'_3} \leq th) \bigr )\lor \bigl ( (th < mp_{t\!f'_3})\land (mp_{t\!f_3}\leq th) \bigr )
\label{equ:samedecisiontwo}
\end{equation}

\noindent Setting \(\varphi_1\equiv (th < mp_{t\!f_3})\land (mp_{t\!f'_3} \leq th)\) and \(\varphi_2 \equiv (th < mp_{t\!f'_3})\land (mp_{t\!f_3}\leq th)\), 
the same theorem tells us that~(\ref{equ:samedecisiontwo}) is equivalent to
\begin{equation}
\bigr [\hbox{not } B^{C_3}_{X_3}\union C B^{C'_3}_{X'_3}\maysat \varphi_1
\bigr ] \hbox{ and } \bigl [\hbox{not }  B^{C_3}_{X_3}\union C B^{C'_3}_{X'_3}\maysat \varphi_2 \bigr ]
\label{equ:samedecisionthree}
\end{equation}

\noindent Using our tool, we determine that
$\mayred{B^{C_3}_{X_3}\union C B^{C'_3}_{X'_3}}{\varphi_1}$ is
unsatisfiable and so~--~by appeal to
Theorem~\ref{theorem:reduction}~--~the first proof obligation
of~(\ref{equ:samedecisionthree}) holds.  Similarly, we evaluate the
satisfiability of $\mayred{B^{C_3}_{X_3}\union C
  B^{C'_3}_{X'_3}}{\varphi_2}$. Our tool reports this to be
satisfiable and so the two constrained BNs do not always support the
same decision. We now want to utilize our non-linear optimization
method to compute ranges of the $th$ itself for which both models
render the same decision. Understanding such a range will be useful to
a modeller as both models are then discovered to be in agreement for
all values of $th$ in such a range.

Since $\mayred{B^{C_3}_{X_3}\union C B^{C'_3}_{X'_3}}{\varphi_1}$ is
unsatisfiable, we use
$C' = \{0 < th < 1, \varphi_2\}$ which forces truth of $\varphi_2$, and compute
$\Sup{}^{\wrap}(th, 0.00000001, B^{C_3}_{X_3}\union {C'} B^{C'_3}_{X'_3})$ to maximise expression $th$.
This obtains the
interval
\begin{equation}
[l,h] = [0.259147588164,0.259147588909] \nonumber
\end{equation}

\noindent Computing $\Inf{}^{\wrap}(th, 0.00000001, B^{C_3}_{X_3}\union {C'}
B^{C'_3}_{X'_3})$ outputs the interval
\begin{equation}
[\tilde l, \tilde h] = [-9.31322e\!\!-\!\!10,-4.65661e\!\!-\!\!10] \nonumber
\end{equation}

\noindent For the given accuracy $\delta$, the interval $[\tilde l,
\tilde h]$ may be interpreted as $0$. Thus, we can say that for all
$th$ in $[0, 0.259147588909]$ the use of either $B^{C_3}_{X_3}$ or
$B^{C'_3}_{X_3}$ could support different decisions. More
importantly, we now know that both constrained BNs always support the same decision as described above when the value of the threshold $th$ for decision making is greater or equal to $0.2592$, say.

The range of $th$ for which both models can support different decisions may  seem rather large and it may be surprising that it goes down to zero. But this is a function of the chance and capability of finding a tamper in an IB. Intuitively, the models tend to disagree most in situations where the chance of cheating by tampering is highest, when $x = 1$, and thus where authenticating the IB has benefit. Our approach gave a decision maker safe knowledge that any threshold for decision making outside the range $[-9.31322e\!\!-\!\!10,0.259147588909]$ would statistically agree and lead to the same decision regarding finding tampers, irrespective of the initial number of IBs~--~either 10 or 20~--~in the pool. Dependent on the nations involved, and the tolerances for decision making they are willing to set, it could be decided~--~for instance~--~that building only $10$ IBs per inspection would be enshrined in the treaty to avoid unnecessary expense and so forth. This would undoubtedly be an important data-driven decision for diplomats and negotiators to make.

\subsection{Symbolic sensitivity analysis}
\label{section:sensitive}
It is well known that BNs may be sensitive to small changes in
probability values in tables of some nodes. Sensitivity
analyses have therefore been devised as a means for assessing the degree
of such sensitivities and the impact this may have on decision
support. See, e.g.,  the \emph{sensitivity value} defined in~\cite{DBLP:journals/tsmc/Laskey95,DBLP:conf/ifip11-9/KwanOCTLL11}.

We now  leverage such analyses to our
approach by computing such sensitivity measures \emph{symbolically} as terms of the
logic $\mathcal Q$. Then we may analyze such terms using the methods
$\Sup{}^{\wrap}$ and $\Inf{}^{\wrap}$ as before to understand how such sensitivity
measures may vary across concretizations of a constrained BN. We
illustrate this capability for 
constrained BN $B^{C_4}_{X_4}$,
which is similar to $B^{C_1}_{X_1}$ but has
probability table for node {\sf Cheating Method} as shown in
Figure~\ref{fig:cBNsix}.

The sensitivity value describes the change in the posterior
output of the hypothesis for small variations in the likelihood of the
evidence under study. The larger the sensitivity value, the less
robust the posterior output of the hypothesis. In other words, a
likelihood value with a large sensitivity value is prone to generate
an inaccurate posterior output. If the sensitivity value is less than
$1$, 
then a small change in the likelihood value has a minimal effect on the result of the posterior output of the hypothesis.
\begin{figure}
\centering
\begin{tabular}{@{}lcc@{}}
\toprule
\multicolumn{3}{c}{{\sf Cheating Method}}                           \\ \midrule
                              & Is cheating & Is not cheating \\
None                          & 0           & 1               \\
IB tamper only                & $x$         & 0               \\
Surrogate source only         & $0.6666-x$         & 0               \\
IB tamper \& surrogate source & 0.3334         & 0               \\ \bottomrule
\end{tabular}
\caption{Probability table for node {\sf Cheating Method} in constrained BN $B^{C_4}_{X_4}$ where $C_4$ contains $\{ 0 < x <
  0.6666, 0 + x + (0.6666 - x) + 0.3334 = 1.0   \}$, $\varx = \{x\}$,
  $\varmp = \{mp_{t\!f}\}$, $X_4 = \varx\cup \varmp$, and where the BN
  graph is that of Figure~\ref{fig:inmmmodel}. All other symbolic
  probability tables for $B^{C_4}_{X_4}$ are as for the BN in Figure~\ref{fig:inmmmodel}\label{fig:cBNsix}}
\end{figure}

A modeller may be uncertain about the sensitivity of event
{\sf Will tamper be found?} = Yes to the authentication of IBs if probabilities
in node {\sf Authentication Capability} of the IB were to change. 
Our tool can compute such a \emph{sensitivity value} $s$
symbolically for the sensitivity of event {\sf Will tamper be found?}
= Yes to small perturbations in probabilities of node {\sf Authentication Capability}. 

The sensitivity value \cite{DBLP:journals/tsmc/Laskey95,DBLP:conf/ifip11-9/KwanOCTLL11} is defined in this instance as
\begin{equation}
\label{equ:sensitivity}
s = \frac{PO*(1-POx)*Px}{(PO*PxO+(1-POx)*Px)^2}
\end{equation}

\noindent where terms $PO, Px, POx$ and $PxO$
are defined as
\begin{eqnarray}
PO &\equiv& p(\textnormal{{\sf Authentication Capability} = Low}) \label{equ:PO}\\
Px &\equiv& p(\textnormal{{\sf Finding a tamper in IB} = Yes}) \nonumber \\
POx &\equiv& p(\textnormal{{\sf Authentication Capability} = Low} \mid
\textnormal{{\sf Finding a tamper in IB} = Yes}) \nonumber \\
PxO &\equiv& p(\textnormal{{\sf Finding a tamper in IB} = Yes} \mid
\textnormal{{\sf Authentication Capability} = Low}) \nonumber 
\end{eqnarray}

\noindent All three marginals of {\sf Authentication Capability}
are considered using just two functions $POx$ and
$1-POx$. In~(\ref{equ:PO}),
$PO$ is a modelling choice that combines the states of Medium and High into
one state ($1-POx$). Term $1-POx$ accounts for situations in which an
inspector is relatively good at authentication, with $POx$
representing situations in which they are less capable. Other modelling choices would lead to a marginally small difference in $s$.  

Our tool can compute an explicit function of $s$ in variable $x$, as
 defined in~(\ref{equ:sensitivity}). This symbolic expression for $s$ is depicted in
Figure \ref{fig:sensitivity} and shown as a function of $x$ in
Figure~\ref{fig:sofx}. This confirms that as the value of
$x$ increases, and thus the probability of  ``IB tamper only'' seen in
Figure \ref{fig:cBNsix} decreases, the marginal of interest for {\sf Will tamper be found?} = Yes becomes less sensitive to
changes in the probabilities of the node {\sf Authentication Capability} of IB. 
\begin{figure}
{\small
\begin{verbatim}
s = 250.0*(0.000333567254313619*x + 0.171472799414097)*
(0.000400681386562896*x + 0.205973332629545)**2*
(0.000400681386562899*x + 0.205973332629545)*
(0.00133627242418726*x + 0.686921064319534)/
((0.19713762029366*x + 0.0638269490499437)*
(4.01363933844916e-6*x**2 + 0.00412648402564936*x + 1.06062534386303)**2)
\end{verbatim}
} 
\caption{Symbolic sensitivity value $s$ in constrained BN
  $B^{C_4}_{X_4}$ for node {\sf Will tamper be found?} in IB with respect
  to event {\sf Authentication Capability}, as a function of sole
  variable $x$ in $\varx$ 
\label{fig:sensitivity}}
\end{figure}
\begin{figure}
\begin{center}
\includegraphics[scale=0.6]{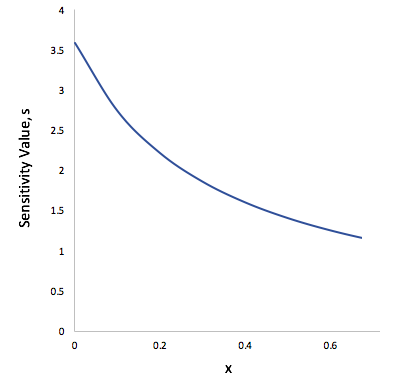}
\end{center}

\caption{Sensitivity value $s$ of Figure~\ref{fig:sensitivity} and~(\ref{equ:sensitivity}) as a
  function of $x$ \label{fig:sofx}}
\end{figure}

We can now determine the \emph{worst-case} sensitivity
value by computing the interval returned by function call
$\Sup{}^{\wrap}(t_s, 0.00000001, B^{C_4}_{X_4})$ as
$[3.5838265468,3.5838265475]$ where $t_s$ is the
term in the righthand side of the equation in
Figure~\ref{fig:sensitivity} that describes $s$ as a function of
$x$. Thus we learn that this
sensitivity cannot be larger than 3.5838265475 for
\emph{all} concretizations of constrained BN $B^{C_4}_{X_4}$. 
As is evident from the graph, there are no valid values of $x$ where
the sensitivity value drops below $1.0$~--~the aforementioned bound at
which a sensitivity score, and therefore its corresponding marginal
probability, is deemed to be robust. 

The output $[1.17313380051,1.17313380116]$ of
$\Inf{}^{\wrap}(s, 0.00000001, B^{C_4}_{X_4})$ confirms this, and shows that $s$ is
always above $1.17313380051$ for \emph{all} concretizations of
constrained BN $B^{C_4}_{X_4}$. Knowing this may indicate to a
decision-maker that potential deviations in the real domain 
from the model of node {\sf Authentication
  Capability} will require close attention, irrespective of the value
of $x$ and thus of the perceived marginal probabilities of the states
of node {\sf Cheating Method}.

\section{Implementation and Evaluation}
\label{section:tool}

\subsection{Software Engineering}
The numerical results reported in previous sections were computed by a
prototype implementation of the approach developed in this paper. This
implementation uses Python to capture a data model for Bayesian
Networks and constraints, to formulate marginals of interest, and to
interface with the SMT solver Z3
\cite{DeMoura:2008:ZES:1792734.1792766}.
The latter we use as a decision procedure for logic $\mathcal Q$ that
also returns witness information for all variables. The computation of
symbolic meaning of marginals relies on the Junction Tree Algorithm
and is achieved through software from an open-source Python package
provided in \cite{ebay14}. 

Python also supports a lightweight and open-source library for
symbolic computation, sympy
\cite{DBLP:journals/peerjpre/MeurerSPCRK0MSR16}, which we can employ 
to run the Junction Tree Algorithm in \cite{ebay14} fully
symbolically. The generated symbolic expressions are then simplified
using a method of sympy before they are put into constraints such as
in~(\ref{equ:mpH}) and added to the SMT solver for analysis.

\subsection{Validation and Evaluation}
Some symbolic marginals that we generated for analyses but not reported in
our case study were too large to be handled by the SMT solver we
used: the string representation of the symbolic meaning was about 25
Megabytes. We performed linear regression on those symbolic expressions and
then validated that this approximation has higher precision than the
accuracy $\delta$, before defining the meaning of marginal variables
as these regressed expressions. Our open-access research data,
discussed on page~\pageref{page:URL}, contains
details on these analyses.

We evaluated the performance of the symbolic interpretation of
the JTA as implemented in \cite{ebay14} on randomly generated
constrained BNs. This does not evaluate
our approach per se, but the manner in which we interpreted an
existing inference implementation symbolically.
We refer to our open-access data repository for
more details on model generation: key parameters are the
number of nodes $\size N$, the number of variables $\size\varx$, and a
random choice of the number of states for each node (between $1$ and
$10$ uniformly at random). Terms in probability tables have form $c$,
$x$ or $1-x$ for constants $c$ or $x$ in $\varx$. 
In generated models, a random node was
picked to determine hard evidence~--~its first state having
probability $1$. The JTA was run for that hard evidence, and the
time to complete it was recorded.  These automated 
test suites ran on an institutional server with 64 Intel Xeon E5/Core i7
processors, on Ubuntu 14.04. 

Many of these tests terminated very quickly. Though, as the number of
nodes per graph and the number of states per node increased, the
running times increased on some but not all tests.
The size of $\varx$ seemed to have a limited effect, possibly
indicating that the additional overhead of our approach to running the
JTA implementation of \cite{ebay14} symbolically in Python is not huge.

\begin{figure}
\centering
\includegraphics[angle=90,scale=0.5]{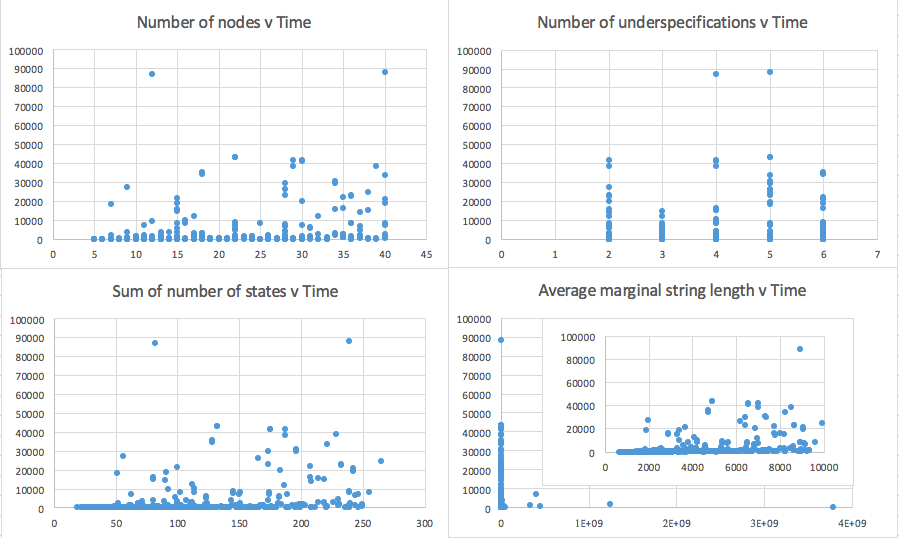}
\caption{Plots for the time the symbolic Junction Tree Algorithm takes
  to run (in seconds, on the $y$-axis) against properties of various randomly generated
  BNs, on the $x$-axis. We assess (from left-to-right, top-to-bottom) the effect on the
  computation time by the number of nodes, the number of variables in $\varx$, the number of states (a summation of the number of states in each of the nodes) and the average length of the text string in the resulting marginal computation. In this last graph, we embed the closeup of the datapoint in the larger graph. \label{fig:auto}}
\end{figure}

Figure~\ref{fig:auto} shows plots for the computation times (in
seconds) of $1000$ such test cases against the number of nodes, the
size of $\varx$, the number of node states in total (a summation over
all nodes) and the average length over all nodes of the outputted
marginal text string in characters. 

For this randomized test suite, there was a small trend for the
running times to increase with the size of the DAG. 
But computations were still quicker for many of the BNs of larger size compared
to smaller ones. The size of $\varx$ appears to have little 
impact on computation time, nor any strong correlation to the
length of the computed symbolic marginal. This suggests that use of
symbolic probabilities may not in and of itself increase such
empirical complexity.

\section{Discussion}
\label{section:discussion}
Our approach advocates the use of constrained Bayesian Networks as a
means of gaining confidence into Bayesian Network modelling and
inference in the face of little or no data. A modeler may thus start
with a BN, turn it into several constrained BNs and subject them to
analysis, and perhaps modify the BN based on such findings.
Witness information computed in analyses could, in principle, be fed back into a BN modeling tool so that users can see a concrete BN that would, for example, explain how a marginal of interest can attain a certain value in a constrained BN.

The ability to represent witness information as a concrete BN is also a means of \emph{testing} whether the computation of symbolic meaning of marginals is free of errors. We have indeed conducted such tests to gain confidence into the correctness of our tool and the packages that it depends upon. Note also that errors in the symbolic meaning of marginals are likely to create numerical inconsistencies, so our analyses would detect such an inconsistent, constrained BN.

The algorithms that we devised for non-linear optimization made no
assumptions about the internal workings of the decision procedure used
and its witness information apart from that such results would be
semantically correct. Knowledge of such internal details could,
however, be exploited to speed up computation. For example, such a
method is used in the SMT solver Z3 to optimize linear objective functions. One could therefore run different methods in parallel or even let them share information in between search iterations.

Our tool prototype interprets the JTA implementation provided in~\cite{ebay14} symbolically, and symbolic
meanings of marginal variables may contain divisions. Of course, we could translate away all division operators without changing meaning~--~to match this with the formal setting of Section~\ref{section:theory}. We did not do this since our foundations apply equally to $\mathcal Q$ extended with division, such translations would increase the size of these terms, and the SMT solver we used, Z3, was able to process and reason with such or suitably simplified terms.

\section{Related Work}
\label{section:related}
In \cite{DBLP:journals/jacm/Darwiche03}, it is shown how probabilistic
inference in Bayesian Networks can be represented through the
evaluation and formal differentiation of a ``network polynomial''. The
size of the polynomial can be reduced by its representation in an
arithmetic circuit, in which evaluation and differentiation are more
efficient. It would be of interest to determine whether this work
can be extended to make the computation of symbolic marginals generated in our
approach more efficient.

For Bayesian Networks there are methods for
learning the structure of a DAG  and for learning the
probabilities within nodes of such a graph (see e.g.\ \cite{heckerman96,cussens15})~--~based on
existing empirical data. We assumed in this paper that little or no data are
available, ruling out the effective use of such learning methods. But
our approach is consistent with settings in which plenty of data are
available. 

Bayesian Networks have tool support such as the software JavaBayes
\cite{javabayes98}, which is able to perform robustness analysis. But
this software can neither 
cope with the Knightian uncertainty of our approach, nor fuse networks
of different structures together with non-trivial constraints.

Our work in \cite{DBLP:conf/esorics/BeaumontEHP15} reported early
attempts of developing the approach presented in this paper: in
\cite{DBLP:conf/esorics/BeaumontEHP15}, a much simpler Bayesian Network of a nuclear
inspection process is presented and some analyses with preliminary
versions of our tool are discussed; but that work offered neither
formal foundations nor greater technical details for the methods it
used. The more detailed Bayesian Network we studied in
Section~\ref{section:casestudy} was discussed in \cite{INMMpaper},
along with a non-technical summary of our general approach and some of its analysis findings.

\emph{Credal networks}~--~see e.g.\ \cite{DBLP:journals/ai/Cozman00}~--~refer
to the theory and
practice of associating a convex set of probability measures with
directed, acyclic graphs. Credal networks 
are also referred to as the Theory of Imprecise Probabilities
\cite{walley91} or as the Quasi-Bayesian Theory \cite{Giron1980}.

The generalization of probability measures to sets of such measures can accommodate a formal notion of probabilistic independence, rooted in axioms of preferences as developed in \cite{DBLP:journals/ai/Cozman00}.
The approach is based
on \emph{constraints} for such convex sets of probability measures.
Inference algorithms and their approximations
are bespoke for an interpretation of constraints; an interpretation is
called an ``extension''
in~\cite{DBLP:journals/ai/Cozman00}. 

To compare this to our approach, we
follow Good's black box model in that our semantics and optimizations reflect Bayesian inference~--~even though this is done symbolically.
Another difference is that a constrained Bayesian Network may have nodes with \emph{non-convex} 
sets of probability measures as meaning, for example
when logical constraints on variables rule out certain points in
intervals. Our approach is also more practical in outlook, since we rely on
reductions to known and tried techniques, such as satisfiability
checking for the existential theory of the reals.
In contrast,
theoretical results for Credal
Networks range from different evidence propagation and inference
methods (see e.g.\ \cite{CozmanRocha02, CozmanRocha02a}) to deep relationships to
logic programming and its semantics
\cite{cozmanmaua16}.

In \cite{boltbock}, a methodology is developed for assessing
sensitivity of lower and upper probabilities in Credal networks. It is
shown that for some classes of parameters in Bayesian networks one may
replace the Credal sets of probability measures associated with such
parameters with a sole such measure. It would be of interest to
determine whether these or similar results are attainable for suitable
classes of constrained Bayesian Networks.

Constraint Networks \cite{dechter} are graphical representations that
are meant to guide solution strategies for constraint satisfaction
problems. In our tool prototype, we decoupled the choice of graph structure for
a constrained Bayesian Network from the use of strategies for solving
satisfiability problems over the existential theory of the reals. It
may be beneficial to couple graph structure and satisfiability
checking in tool support of our approach that relies on constraint
satisfaction solvers.

\section{Conclusions}
\label{section:conclusions}
This work was motivated by the fact that some problem domains have little or no data that one could use to learn the structure of a causal network or the probabilities for nodes within that structure~--~whatever the reasons for such sparsity of data may be in such a domain.  This led us to consider suitable generalizations of Bayesian Networks. Ideally, we wanted a formalism that those who already use Bayesian Networks for modeling and analysis would find easy to adopt. In particular, we sought to preserve~--~as much as possible~--~the manner in which probabilistic inference is done in Bayesian Networks. Crucially, we wanted a set of methods whose use could help us to build sufficient confidence into the quality, suitability or robustness of models expressed in such a formalisms in the face of little or no empirical data.

We propose constrained Bayesian Networks as such a formalism. The derivation of that concept is a contribution in and of itself, and it used first-order logic
and its semantics as well as syntactic criteria for wellformedness. But it also required methods from three-valued logic to
define a precise yet intuitive \emph{semantics} for a constrained
BN. 

We also developed meta-properties of this semantics, including checks for the consistency of a constrained Bayesian Network. These properties were needed to prove the correctness of our optimization algorithms, which can compute suprema or infima of bounded arithmetic terms up to a specific accuracy. These optimization algorithms are non-standard in that they rely on a decision procedure for the theory of reals and in that the optimization problems are generally non-linear and non-convex.

The marginals in a constrained Bayesian Network are computed
\emph{symbolically}, but computed in the same manner as the marginals
for a Bayesian Network~--~a concretization of that constrained
Bayesian Network. This is appealing as it allows reuse of known and
trusted methods such as the Junction Tree Algorithm. But it also
creates a potential computational bottleneck with scope for future
work that may extend an approach in
\cite{DBLP:journals/jacm/Darwiche03} to our setting.

We implemented our approach in a tool prototype, which benefitted from
the significant advances in symbolic computation
and in the implementation of theorem
provers such as SMT solvers. We evaluated this prototype through
stress tests and
a non-trivial case study in the domain of nuclear arms
control. The latter is a domain in which the availability of data is very limited and where any means of building confidence into the trustworthiness of mathematical models are expected to have positive impact on arms reduction efforts.

 We used this case study to illustrate some pertinent types of
 analyses of a constrained Bayesian Network that our approach can
 accommodate: a \emph{range analysis} that computes infima and suprema
 for a term of interest to determine their robustness, the comparison of two or more constrained Bayesian Networks to assess modeling impact, the ability to determine ranges of threshold values that would render equivalent decision support, and the symbolic computation of a sensitivity measure for a given node~--~with the ability to optimize this to understand worst-case sensitivities.
We trust that the approach presented in this paper will be useful for other applications in the arms-control domain, as well as in other domains~--~particularly those with a lack of data.

\paragraph{{\bf Acknowledgements:}} This work was supported by
AWE plc, and in part by the UK Engineering and Physical Sciences Research
Council grants EP/N020030/1 and EP/N023242/1.

\paragraph{{\bf Open Access:}}  The Python and SMT code for the queries and
models of this paper and raw SMT analysis results are found in the
public data repository\newline
\begin{center}
\verb+bitbucket.org/pjbeaumont/beaumonthuthcbns/+ \label{page:URL}
\end{center}

\appendix
\section{Mathematical Proofs}
\beginproof{Theorem~\ref{theorem:dual}}
\begin{enumerate}
\item We have that $B^C_X\mustsat \phi$ holds iff for all concretizations
  $B^C_X[\alpha]$ of $B^C_X$ we have that
  $\alpha\models\phi$ holds iff for all concretizations
  $B^C_X[\alpha]$ of $B^C_X$ we have that $\alpha\models
  \lnot\phi$ does not hold iff 
  $B^C_X\maysat\lnot\phi$ does not hold.

\item We have that $B^C_X\maysat \phi$ holds iff there is some concretization
  $B^C_X[\alpha]$ of $B^C_X$ such that 
  $\alpha\models\phi$ holds iff there is some concretization
  $B^C_X[\alpha]$ of $B^C_X$ such that 
  $\alpha\models\neg\phi$ does not hold iff 
  $B^C_X\mustsat\lnot\phi$ does not hold.

\item 
  \begin{enumerate}
  \item Let $B^C_X\mustsat\phi_1\land\phi_2$ hold. Let $B^C_X[\alpha]$
    be a concretization of $B^C_X$. Then we know that
    $\alpha\models \phi_1\land \phi_2$ holds. This
    implies that $\alpha\models\phi_i$ holds for
    $i=1,2$. But then both $B^C_X\mustsat\phi_1$ and
    $B^C_X\mustsat\phi_2$ hold since $B^C_X[\alpha]$ was an
    arbitrary concretization of $B^C_X$.
 
  \item Let both $B^C_X\mustsat\phi_1$ and
    $B^C_X\mustsat\phi_2$ hold. Let $B^C_X[\alpha]$ be a
    concretization of $B^C_X$. Then $B^C_X\mustsat\phi_i$ implies
    that $\alpha\models \phi_i$ holds for
    $i=1,2$. Therefore, we get that $\alpha\models
    \phi_1\land\phi_2$ holds as well.
    Since $B^C_X[\alpha]$ was an arbitrary concretization of
    $B^C_X$, this gives us that  $B^C_X\mustsat\phi_1\land\phi_2$ holds.

  \end{enumerate}

\item 
  \begin{enumerate}
  \item Let $B^C_X\maysat\phi_1\lor\phi_2$ hold. Then there is
    some concretization $B^C_X[\alpha]$
   of $B^C_X$ such that 
   $\alpha\models \phi_1\lor \phi_2$ holds. This
    implies that $\alpha\models \phi_i$ holds for
    \emph{some} $i=1,2$. But then $B^C_X\maysat\phi_i$ holds
    as claimed.
 
  \item Let one of $B^C_X\maysat\phi_1$ and
    $B^C_X\maysat\phi_2$ hold, say $B^C_X\maysat\phi_i$. 
   Then there is some concretization $B^C_X[\alpha]$ of $B^C_X$ such
   that $\alpha\models\phi_i$ holds. This implies
   that
 $\alpha\models
    \phi_1\lor\phi_2$ holds as well.
    Since $B^C_X[\alpha]$ is a concretization of
    $B^C_X$, we get that  $B^C_X\maysat\phi_1\lor\phi_2$ holds.

  \end{enumerate}
\end{enumerate}

\QED

\beginproof{Theorem~\ref{theorem:conchar}}
\begin{itemize}
\item \emph{Item~1 implies item~2:}
 Let $B^C_X\maysat\true$ hold. By definition of $\maysat$, there
  then is some concretization $B^C_X[\alpha]$ of $B^C_X$ such that
  $\alpha\models \true$ holds. Therefore the set of
  concretizations of $B^C_X$ is non-empty and so $B^C_X$ is
  consistent.

\item \emph{Item~2 implies item~3:} 
Let $B^C_X$ be consistent. Suppose that $\phi$ is in $\mathcal
  Q$ such that $B^C_X\mustsat\phi$ holds. Since $B^C_X$ is consistent,
  there is some concretization $B^C_X[\alpha]$ of $B^C_X$. Since
  $B^C_X\mustsat\phi$ holds, we get that $\alpha\models
  \phi$ holds. But then we have $B^C_X\maysat \phi$ be definition of
  $\maysat$.

\item \emph{Item~3 implies item~4:}  
Let $B^C_X\mustsat\phi$ imply $B^C_X\maysat \phi$ for all $\phi$
  in $\mathcal Q$. Let $\psi$ be in $\mathcal Q$. We claim that
  $B^C_X\maysat \psi\lor\lnot\psi$ holds. By
  Theorem~\ref{theorem:dual}.4, it suffices to show that $B^C_X\maysat
  \psi$ or $B^C_X\maysat\lnot\psi$ holds. If the former holds, we are
  done. Otherwise, we have that $B^C_X\maysat
  \psi$ does not hold. By Theorem~\ref{theorem:dual}.2, this implies that
  $B^C_X\mustsat \lnot\psi$ holds. 

\begin{itemize}
\item Next, we show that $B^C_X$ has to be consistent: note that
$B^C_X\mustsat\true$ holds by the definitions of $\mustsat$ and
$\models$. Therefore, we get that $B^C_X\maysat\true$ holds by item~3~--~and we already
showed that this implies that $B^C_X$ is consistent. 
\end{itemize}

Let $B^C_X[\alpha]$ be a concretization of
  $B^C_X$, which exists as $B^C_X$ is consistent. Since we showed $B^C_X\mustsat \lnot\phi$, 
the latter implies that $\alpha\models\lnot\phi$. But
then $B^C_X\maysat\lnot\phi$ follows given the definition of
$\maysat$.

\item \emph{Item~4 implies item~5:}    
Let $B^C_X\maysat \phi\lor\lnot\phi$ hold for all $\phi$ in
  $\mathcal Q$. Let $\psi$ be in $\mathcal Q$. Then we have that
$B^C_X\maysat \psi\lor\lnot\psi$ holds, and we need to show that
  $B^C_X\mustsat \psi\land\lnot\psi$ does not hold. Proof by
  contradiction: assume that $B^C_X\mustsat \psi\land\lnot\psi$
  holds. Since $B^C_X\maysat \psi\lor\lnot\psi$ holds, we know
that there is some concretization
  $B^C_X[\alpha]$ of $B^C_X$ such that $\alpha\models
  \psi\lor\lnot\psi$ holds.  Since $B^C_X\mustsat
  \psi\land\lnot\psi$ holds, we know that $B^C_X\mustsat \psi$ and
  $B^C_X\mustsat\lnot\psi$ hold by Theorem~\ref{theorem:dual}.3. 
We do a case analysis on the truth of judgment $\alpha\models
  \psi\lor\lnot\psi$:

\begin{itemize}
\item Let $\alpha\models \psi$ hold . Since
  $B^C_X\mustsat\lnot\psi$ holds, this
implies that $\alpha\models\lnot\psi$ holds. This
  contradicts that $\alpha\models
  \psi$ holds.

\item Let $\alpha\models\lnot\psi$ hold. Since $B^C_X\mustsat \psi$
  holds, this implies that $\alpha\models\psi$
  holds. This contradicts that $\alpha\models
  \lnot\psi$ holds.
\end{itemize}

\item \emph{Item~5 implies item~1:} Let $B^C_X\mustsat \phi\land\lnot\phi$ not hold, for all $\phi$ in $\mathcal Q$. Since $\true$ is in $\mathcal Q$,
  we know that $B^C_X\mustsat\true\land\lnot\true$ does not hold. By
  definition of $\mustsat$ and $\models$, we have that
  $B^C_X\mustsat\true$ holds. By Theorem~\ref{theorem:dual}.3 and
  since $B^C_X\mustsat\true\land\lnot\true$ does not hold, we infer
  that $B^C_X\mustsat\lnot\true$ does not hold. By
  Theorem~\ref{theorem:dual}.2, this implies that $B^C_X\maysat\true$
  holds as claimed.
\end{itemize}
\QED

\beginproof{Theorem~\ref{theorem:reduction}}
\begin{enumerate}
\item Constraints $\varphi'$ in $C$ and $\varphi$ are quantifier-free formulas of $\mathcal Q$ with variables contained in $X$, which equals $\{x_1,x_2,\dots, x_n\}$. Therefore, the formula in~(\ref{equ:maysatreduction}) is in $\mathcal Q$, and contains only existential quantifiers and all in front of the formula.

\item We prove this claim by structural induction over $\varphi$:
\begin{itemize}
\item Let $\varphi$ be $\true$. Then $\mayred{B^C_X}{\true}$ equals
  $\exists x_1\colon \dots \colon \exists x_n\colon \true\land
  \bigwedge_{\varphi'\in C}\varphi'$ and this is satisfiable iff there
  is an assignment $\alpha$ such that $\alpha\models
  \bigwedge_{\varphi'\in C} \varphi'$ and $\alpha\models\true$ both hold (the latter holding by definition) iff there is a concretization $B^C_X[\alpha]$ of $B^C_X$ iff $B^C_X$ is consistent iff (by Theorem~\ref{theorem:conchar}) $B^C_X\maysat \true$ holds. 

\item Let $\varphi$ be $t_1\leq t_2$. Then $\mayred{B^C_X}{t_1\leq
    t_2}$ equals $\exists x_1\colon \dots \colon \exists x_n\colon
  (t_1\leq t_2)\land \bigwedge_{\varphi'\in C}\varphi'$ and this is
  satisfiable iff there is an assignment $\alpha$ such that
  $\alpha\models \bigwedge_{\varphi'\in C} \varphi'$ and $\alpha\models t_1\leq t_2$ both hold iff there is a concretization $B^C_X[\alpha]$ of $B^C_X$ such that $\alpha\models t_1\leq t_2$ holds iff $B^C_X\maysat t_1\leq t_2$ holds.

\item Let $\varphi$ be $t_1 < t_2$. Then $\mayred{B^C_X}{t_1< t_2}$
  equals $\exists x_1\colon \dots \colon \exists x_n\colon (t_1<
  t_2)\land \bigwedge_{\varphi'\in C}\varphi'$ and this is satisfiable
  iff there is an assignment $\alpha$ such that $\alpha\models
  \bigwedge_{\varphi'\in C} \varphi'$ and $\alpha\models t_1< t_2$ both hold iff there is a concretization $B^C_X[\alpha]$ of $B^C_X$ such that $\alpha\models t_1< t_2$ holds iff $B^C_X\maysat t_1< t_2$ holds.

\item Let $\varphi$ be $\lnot\psi$. Then $\mayred{B^C_X}{\lnot\psi}$
  equals $\exists x_1\colon \dots \colon \exists x_n\colon \lnot\psi
  \land \bigwedge_{\varphi'\in C}\varphi'$ and this is satisfiable iff
  there is an assignment $\alpha$ such that $\alpha\models
  \bigwedge_{\varphi'\in C} \varphi'$ and $\alpha\models \lnot\psi$ both hold iff there is a concretization $B^C_X[\alpha]$ of $B^C_X$ such that $\alpha\models\lnot\psi$ holds iff $B^C_X\maysat \lnot\psi$ holds.

\item Let $\varphi$ be $\varphi_1\land \varphi_2$. Then
  $\mayred{B^C_X}{\varphi_1\land\varphi_2}$ equals $\exists x_1\colon
  \dots \colon \exists x_n\colon \varphi_1\land\varphi_2 \land
  \bigwedge_{\varphi'\in C}\varphi'$ and this is satisfiable iff there
  is an assignment $\alpha$ such that $\alpha\models
  \bigwedge_{\varphi'\in C} \varphi'$ and $\alpha\models \varphi_1\land\varphi_2$ both hold  iff there is a concretization $B^C_X[\alpha]$ of $B^C_X$ such that $\alpha\models\varphi_1\land\varphi_2$ holds iff $B^C_X\maysat \varphi_1\land\varphi_2$ holds.
\end{itemize}

\item By the previous item, we may decide $B^C_X\maysat\varphi$ by deciding whether formula $\mayred{B^C_X}\varphi$ is satisfiable. By item~1 above, that formula is in the existential fragment of $\mathcal Q$. By \cite{Canny88}, deciding the satisfiability (truth) of such formulas is in PSPACE in the size of such formulas.

\item By Theorem~\ref{theorem:dual}.1, we have that $B^C_X\mustsat \varphi$ holds iff $B^C_X\maysat \lnot\varphi$ does not hold. By item~2 above, the latter is equivalent to $\mayred{B^C_X}{\lnot \varphi}$ not being satisfiable. By \cite{Canny88}, this can be decided in PSPACE in the size of formula $\mayred{B^C_X}{\lnot \varphi}$.
\end{enumerate}
\QED

\beginproof{Theorem~\ref{theorem:apprsup}}
The arguments below make use of Theorems~\ref{theorem:dual}
and~\ref{theorem:reduction} without explicit reference to them.
Note that consistency of $B^C_X$ and 
$0 < \sup\tval t$ guarantee that the first $let$ statement
in $\Sup$ can
find such a $\alpha$. In particular, we see that $0 < cache$ becomes
an invariant and so $cache < 2*cache$ is another invariant.

\begin{enumerate}
\item First, we show that the $asserts$ hold prior to the execution of the
second $while$ loop. Note that $cache$ is always assigned reals of
form $\means t\eta$ for
some concretization $B^C_X[\eta]$ of $B^C_X$. So when $low$ is
initialized with the last updated value of $cache$, then $B^C_X\maysat
t\leq low$ clearly holds after the first assignment to $low$ 
(witnessed by the assignment that gave rise to the
last value of $cache$) and prior
to its reassignment.  By definition of the initial value of $high$,
we have that $B^C_X\maysat t\geq high$ does not hold after that
initial assignment and prior to the reassignment of $high$. Therefore,
both $asserts$ in front of the second $while$ loop hold, and we get
that $low\leq high$ is an invariant.

Second, we show that each iteration of the second $while$ loop
preserves the $asserts$. This is clear as the Boolean guard of the
$i\!f$ statement tests for preservation of these $asserts$, and makes
the correct, invariant-preserving assignment accordingly.

Third, let $[l,h]$ be the returned closed interval. It is clear that
$h-l\leq \delta$ holds as required. We argue that
$\sup\tval t$ is in $[l,h]$. Since the $asserts$ hold for $l$ and
$h$, we know that $B^C_X\maysat t \geq l$ holds, but $B^C_X\maysat
t\geq h$ does not hold. Let $c$ be in $\tval t$. Then there is some
$\alpha$ with $c = \means t\alpha$. Since $B^C_X\maysat
t\geq h$ does not hold, we get that $\means t\alpha < \means h\alpha =
h$. Therefore, $h$ is an upper bound of $\tval t$ which implies
$\sup\tval t\leq h$. Since
$B^C_X\maysat t \geq l$ holds, we have some concretization
$B^C_X[\alpha']$ with $\alpha'\models t\geq l$. This
means $\means t{\alpha'} \geq l$. But $\sup\tval t\geq \means
t{\alpha'}$ as the latter is an element of $\tval t$. Thus, $l\leq
\sup\tval t$ follows.

\item Let $s$ be $\sup\tval t$. For the first $while$ loop, we have at
  least $k$ iterations if $s\geq 2^k\cdot c$, i.e.\ if $s\cdot c^{-1}
  \geq k$, i.e.\ if $k\leq \log_2(s) - \log_2(c)$. So the real number 
$\log_2(s) - \log_2(c)$ is an upper bound on the number of iterations
of the first $while$ loop.

To get an upper bound for the number of iterations of the second
$while$ loop, we know that $high$ is of form $2^{l+1}\cdot c$ and so
$low$ equals $2^l\cdot c$. But then $high-low$ equals $2^l\cdot
c$. Since this is monotone in $l$, we may use the upper bound for the
number of iterations of the first $while$ loop as an upper bound of
$l$, to get $2^{\log_2(s)-\log_2(c)}\cdot c = s\cdot c^{-1}\cdot c =
s$ as an upper bound on the value of $\mid high-low \mid$ before the
Boolean guard of the second $while$ is first evaluated. This allows us
to derive an upper bound on the number of iterations of the second
$while$ loop, since the larger that value is, the more iterations take
place. Based on the bisection in each iteration, there are at least
$k$ iterations if $s\cdot 2^{-k} > \delta$, i.e.\ if $k < \log_2(s) -
\log_2(\delta)$. Therefore, the total number of iterations of both
$while$ loops combined is $(\log_2(s) - \log_2(c)) + (\log_2(s) -
\log_2(\delta))$. The claim now follows given that each iteration
makes exactly one satisfiability check and since there is an initial
satisfiability check as well.
\end{enumerate}
\QED

\beginproof{Theorem~\ref{theorem:apprinf}}
\begin{enumerate}
\item The argument is similar to the one for
Theorem~\ref{theorem:apprsup} but there are important differences. 
Note that $cache > 0$ is also here an invariant, 
guaranteed by the fact that $\tval t$ contains a positive
real. We know
that $(0.5^n*cache)_{n\in\Nat}$ converges to $0$ for any positive constant
$cache$. Since $0 < \delta$ and since $\means t{\alpha'}\leq 0.5*cache$ for
the $\means t{\alpha'}$ assigned to $cache$, there is some $n_0$ such that
$0.5^{n_0}*cache \leq \delta$. This proves
that the first $while$ statement terminates. 

\begin{enumerate}
\item Suppose that the return statement in the line after the first $while$
loop is executed. Then $\mayred{B^C_X}{t \leq 0.5*cache}$ is
satisfiable and so there is some concretization $B^C_X[\alpha]$ such that $\means t\alpha
\leq 0.5*cache$. But then $\inf\tval t\leq 0.5*cache$ as well. From
$\tval t\subseteq \Reals_0^+$, we get $0\leq \inf\tval
t$. Therefore, $\inf\tval t$ is in the returned interval
$[0,0.5*cache]$ and $B^C_X\maysat t\leq 0.5*cache$ is true. Moreover,
the length of the interval is $0.5*cache$, which must be less than or
equal to $\delta$ as the first $while$ loop just terminated and the
first conjunct of its Boolean guard is true~--~forcing $0.5*cache >
\delta$ to be false.

\item Otherwise, $\mayred{B^C_X}{t \leq 0.5*cache}$ is not
satisfiable but  the formula $\mayred{B^C_X}{t \leq cache}$ is satisfiable.
From that, it should then be clear
that the $asserts$ in front of the second $while$ statement hold
when they are reached. That each iteration of the second $while$
statement  maintains these two $asserts$ is reasoned similarly as for
$\Sup$.  

So we have that $B^C_X\maysat t\leq h$ and $B^C_X\mustsat t >
l$ are invariants. 
This means that $l$ is a lower bound of $\tval t$ and $\means t\alpha\leq h$
for some $\means t\alpha$ in $\tval t$. But then $l\leq \inf\tval
t\leq \means t\alpha\leq h$
shows that $\inf\tval t$ is in $[l,h]$. 
\end{enumerate}

\item Let $i$ be $\inf\tval t$. We derive an upper bound on the number
  of iterations for the first $while$ loop. Because we are interested
  in upper bounds, we may assume that the $\alpha'(t)$ assigned to
  $cache$ equals $0.5\cdot cache$ for the current value of $cache$. We
  then have at least $k$ iterations if $\delta < c\cdot 2^{-k}$ and
  $i\leq c\cdot 2^{-k}$. Since we are interested in upper bounds on
  that number of iterations, we get at least $k$ iterations if both $\delta \leq c\cdot 2^{-k}$ and
  $i\leq c\cdot 2^{-k}$ hold, i.e.\ if $\min(\delta,i)\leq c\cdot
  2^{-k}$. But this is equivalent to $k\leq \log_2(c) -
  \log_2(\min(i,\delta))$. 

We now derive an upper bound on the number of iterations of the second
$while$ loop. The initial value of $high-low$ equals $cache - 0.5\cdot
cache = 0.5\cdot cache$ for the current value of $cache$ when entering
that loop. The value of $cache$ is monotonically decreasing during
program execution and so $c/2$ is an upper bound of $high-low$. We may
therefore use $c/2$ as initial value of $high-low$ since this can only
increase the number of iterations, for which we seek an upper
bound. There are now at least $k$ iterations if $(c/2)\cdot 2^{-k} >
\delta$ which is equivalent to $k < \log_2(c) - 1 - \log_2(\delta)$.

The total number of iterations for both $while$ loops is therefore $(\log_2(c) -
  \log_2(\min(i,\delta))) + (\log_2(c) - 1 - \log_2(\delta)) = 2\cdot
  log_2(c) - \log_2(\min(i,\delta)) - 1$. From this the claim follows
  since each iteration has exactly one satisfiability check of the
  stated form, and there is one more satisfiability check between the
  first and second $while$ loop.
\QED
\end{enumerate}

\beginproof{Theorem~\ref{theorem:supwrapper}}
\begin{enumerate}
\item We do a case analysis:

  \begin{enumerate}
  \item If algorithm $\Sup$ is called, then consistency of $B^C_X$
    and $0 < \sup\tval t$ follow from the Boolean guard that triggered the call. Since
  $\sup\tval t < \infty$ is assumed, we get $0 < \sup\tval t <
  \infty$ and so $\Sup$ terminates by Theorem~\ref{theorem:apprsup}.

  \item If $0$ is returned as a maximum, the algorithm clearly
    terminates and no pre-conditions are needed.

  \item If $\Inf$ is called, we have to show that $\tval{-t}$ is
    a subset of $\Reals_0^+$ that contains a positive real. 
   Since the first two $return$ statements were not
    reached, we know that $B^C_X$ is consistent and 
    $\tval t$ is a subset of $\Reals^-$. But
    then $\tval {-t}$ is a subset of $\Reals^+$. 
  \end{enumerate}

\item If the algorithm reports that $0$ is the maximum for $t$, then we know that $\tval t$ cannot contain a positive real (first if-statement), and that it contains $0$ (second if-statement). Clearly, this means that $0$ is the supremum of $\tval t$ and so also its maximum as $0$ is in $\tval t$.

\item Let 
  $\Sup{}^{\wrap}(t,\delta,B^C_X)$ return an interval $[-h,-l]$. Then $[l,h]$ is the interval returned by a call to $\Inf(-t,\delta,B_X^C)$. By the first item and Theorem~\ref{theorem:apprinf}, we get that $B_X^C\maysat -t\leq h$ holds,  $\inf\tval {-t}$ is in $[l,h]$, and $h-l\leq\delta$ . Therefore, we conclude that
$B_X^C\maysat t\geq -h$ holds as claimed. Moreover, since $\inf\tval {-t}$
   equals $-\sup\tval t$, this implies that $\sup\tval t$ is in the
   closed interval $[-h,-l]$, whose length is that of $[l,h]$ and so
   $\leq \delta$.

\item If the algorithm returns saying that $B^C_X$ is inconsistent,
  then all three formulas $\mayred{B^C_X}{t > 0}$, $\mayred{B^C_X}{t
    =0}$, and $\mayred{B^C_X}{t < 0}$ are unsatisfiable. But then we
  know that the three judgments $B^C_X\maysat t > 0$, $B^C_X\maysat t
  = 0$, and $B^C_X\maysat t <  0$ do not hold, by
  Theorem~\ref{theorem:reduction}. 
This means that $B^C_X$
  is inconsistent: for all concretization $B^C_X[\alpha]$ we
  have that $\alpha \models (t > 0)\lor (t=0)\lor (t <
  0)$ holds as that query is a tautology over the theory of reals; and
  then Theorem~\ref{theorem:reduction}.4 yields a contradiction to
  $B^C_X$ being consistent.

\end{enumerate}
\QED

\beginproof{Theorem~\ref{theorem:infimum}}
The correctness of the first two claims in that theorem (inconsistency
and minimum) for $-\infty < \inf\tval t$ follows from the corresponding items of Theorem~\ref{theorem:supwrapper}.
The general identity $\inf
\{x_i\mid i\in I\} = -\sup \{-x_i\mid i\in I\}$ shows that
$-\infty < \inf \{x_i\mid i\in I\}$ iff $-\sup \{-x_i\mid i\in I\} < \infty$ and so preconditions are also met.
Finally, to see the correctness of $\Inf{}^{\wrap}$ when interval $[l,h]$ is returned, note that this means that interval $[-h,-l]$ is returned for the call $\Sup{}^{\wrap}(-t,\delta,B_X^C)$ and so $B_X^C\maysat -t\geq -h$ holds by Theorem~\ref{theorem:supwrapper}. But this implies that $B^C_X\maysat t\leq h$ holds as claimed.
\QED

\section{Quantitative Information about the BN of Figure~\ref{fig:inmmmodel}}
Table~\ref{fig:complex3} shows quantitative information about the size and complexity of the BN in Figure~\ref{fig:inmmmodel}.
\begin{table}[ht]
\centering
\includegraphics[scale=0.625]{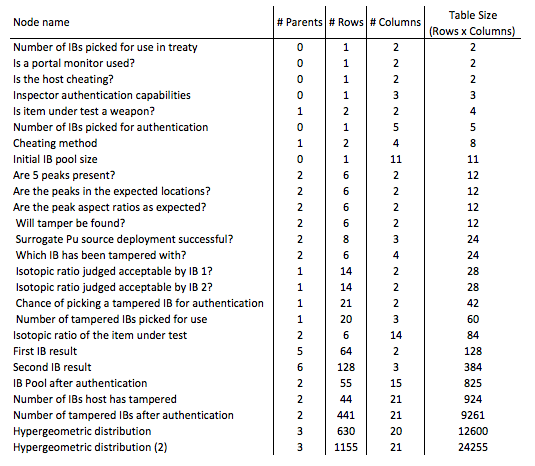}
\caption{The probability tables for the BN in
  Figure \ref{fig:inmmmodel} are too large to be specified explicitly
  in the paper. Here we want to convey the structural and resulting
  computational complexity of
  this BN in tabular form. For each node shown in the leftmost column
  we list its number of parents ({\sf \# Parents}) in the BN graph,
  its number of input combinations ({\sf \# Rows}) which is
  $\max(1,\prod_{i=1}^k n_i )$ where $n_i$ is the number of outputs any
  of the $0\leq k$ parents $i$ can have, whereas in {\sf \# Columns}
  we list the number of output values that node itself can have.
Rightmost column {\sf Table Size} depicts the size of the support of
the probability distribution of that node.
\label{fig:complex3}}
\end{table}

\end{document}